\documentclass[11pt]{scaleai-paper}

\usepackage{amsmath}
\usepackage{amssymb}
\usepackage{multirow}
\usepackage{tabularx}
\usepackage[square,numbers,sort&compress]{natbib}
\usepackage[colorlinks=true,linkcolor=scaleLink,citecolor=scaleLink,urlcolor=scaleLink]{hyperref}
\hypersetup{pdftitle={To Critic Or Not To Critic},pdfauthor={Jason Wei, George Pu, Jingxuan Fan, Yuan (Emily) Xue, Yuan (Christy) Li}}

\papertype{Scale AI Research}

\author[1,2]{Jason Wei\thanks{Work conducted during an internship at Scale AI.}}
\author[1]{George Pu}
\author[1]{Jingxuan Fan}
\author[1]{Yuan (Emily) Xue}
\author[1]{Yuan (Christy) Li}
\affil[1]{Scale AI}
\affil[2]{Carnegie Mellon University}

\makeatletter
\let\scale@xfloat\@xfloat
\def\@xfloat#1[#2]{\scale@xfloat#1[!tp]}

\makeatother

\usepackage{latexsym}
\usepackage[T1]{fontenc}
\usepackage[utf8]{inputenc}
\usepackage{microtype}
\usepackage{inconsolata}

\usepackage{amsmath,amsfonts,bm}

\def\eqref#1{equation~\ref{#1}}

\def\1{\bm{1}}

\DeclareMathAlphabet{\mathsfit}{\encodingdefault}{\sfdefault}{m}{sl}
\SetMathAlphabet{\mathsfit}{bold}{\encodingdefault}{\sfdefault}{bx}{n}

\usepackage{url}
\usepackage{array}
\usepackage{booktabs}
\usepackage{xcolor}
\usepackage{colortbl}
\usepackage{xurl}
\usepackage{graphicx}
\usepackage{amsthm}
\usepackage{enumitem}

\theoremstyle{definition}
\newtheorem{proposition}{Proposition}

\title{To Critic Or Not To Critic}

\begin{document}

\renewcommand{\thefootnote}{\fnsymbol{footnote}}
\setcounter{footnote}{1}
\maketitle
\renewcommand{\thefootnote}{\arabic{footnote}}
\setcounter{footnote}{0}

\begin{abstract}
As large language models tackle longer and more complex tasks, understanding when a critic improves the efficiency of reinforcement learning (RL) becomes increasingly important. Proximal Policy Optimization (PPO) trains a critic to reduce variance and provide fine-grained credit assignment, while Group Relative Policy Optimization (GRPO) drops the critic and avoids the bias of bootstrapping from the critic's value estimates. 
Although recent critic-based methods have reported strong performance relative to GRPO, it remains unclear when each approach is preferable. We argue that there is no universal answer; instead, the choice depends on several interacting factors, including batch size, critic reliance, and task complexity. We study the bias--variance tradeoff in terms of gradient direction, measuring how closely each method's gradient estimates align with the true policy gradient as batch size grows.
Under a shared actor objective, we compare GRPO with PPO in control tasks, short-horizon math, and long-horizon terminal agents.
Across more than 1,200 training runs, including 96 with LLMs, we find that PPO generally leads at small batch sizes, while GRPO often overtakes heavily bootstrapped PPO and reaches parity with lightly bootstrapped PPO at larger batches. At longer horizons and in more complex domains, GRPO needs larger batches to catch up to PPO's performance.
Accounting for training compute, critics pay off at small budgets, but their returns diminish at larger budgets.
\end{abstract}

\section{Introduction}
\label{sec:intro}

Reinforcement learning (RL) is an essential stage of large language model (LLM) post-training. 
Early pipelines for RL from human feedback relied predominantly on Proximal Policy Optimization (PPO)~\citep{schulman2017ppo, ouyang2022instructgpt}, which trains a separate critic network to reduce gradient variance and provide token-level credit assignment through Generalized Advantage Estimation (GAE)~\citep{schulman2015gae}. Since DeepSeek-R1 demonstrated strong reasoning performance with critic-free Group Relative Policy Optimization (GRPO)~\citep{guo2025deepseekr1,shao2024deepseekmath}, the open-source community has largely abandoned critics, citing reduced memory, implementation simplicity, and unreliable value estimation~\citep{shao2024deepseekmath,yuan2025vcppo,kazemnejad2024vineppo,venkatraman2026lecritique}. 

More recently, critic-based methods have reported strong results relative
to GRPO, renewing interest in critics~\citep{hu2025openreasonerzero,yue2025vapo,liu2025asyppo,sao2026,qi2026bestpracticecriticoptimization}. However, these comparisons are typically made at a single batch size and in
a narrow set of domains. For example, VAPO and SAO train with batch sizes of
512 and 128, respectively, in math and coding~\citep{yue2025vapo,sao2026}. It therefore remains unclear
whether these gains hold at other batch sizes or in other domains.

As LLMs tackle longer agentic tasks, individual rollouts grow more
expensive, making the choice of advantage estimator increasingly
consequential~\citep{yuan2025vcppo,deoliveira2025learningwithoutcritics,li2025turnppo}. We argue that neither GRPO nor PPO
is universally preferable. Instead, the choice depends on several
interacting factors, and we focus on three that shape the bias--variance
tradeoff. 

The first is \emph{batch size}, the number of trajectories used per actor
update. Larger batches reduce gradient variance but not bias, so the relative performance of PPO and GRPO can change with batch size. The second
is \emph{critic reliance}. GRPO avoids the critic entirely, while PPO's GAE
parameter $\lambda$ controls how strongly its advantage calculation
bootstraps from the critic. More bootstrapping typically lowers variance at
the cost of bias from critic approximation error~\citep{schulman2015gae}. The third is \emph{task complexity}, which we study through horizon and domain. Longer horizons increase our upper bound on
gradient variance (Section~\ref{sec:horizon-batch-theory}). Domains differ
in properties that may also raise gradient noise, such as the dimensionality
of the state and action spaces and the diversity of trajectories relevant to
learning~\citep{wu2018variance,yin2018gradientdiversity}. We find that both can make variance reduction more valuable
(Sections~\ref{sec:batch-critic-analysis} and~\ref{sec:llm-gradient-results}).

Prior work has studied these factors largely in isolation, including batch size scaling~\citep{li2026largerbatches,cheng2026isocompute,chen2025gpg},
critic reliance in
PPO~\citep{hu2025openreasonerzero,yuan2025vcppo,qi2026bestpracticecriticoptimization},
and the growth of gradient variance with
horizon~\citep{zhao2011analysis,li2026otb}.
However, how they jointly affect the relative performance of PPO and GRPO across
domains remains poorly understood. We address this gap across more than 1,200
training runs, including 96 with LLMs, by comparing the two methods' advantage
estimators under a shared actor objective while varying batch
size, critic reliance, and task complexity (Section~\ref{sec:comparison-protocol}). We study three settings
(Table~\ref{tab:settings}): control tasks with Pendulum and HalfCheetah, short-horizon single-turn math, and
long-horizon multi-turn terminal agents. Our
contributions are as follows:
\begin{enumerate}[itemsep=0pt]
    \item \textbf{Batch size, critic reliance, and task complexity affect when GRPO catches up.}
    PPO generally leads at small batch sizes, while GRPO often overtakes heavily bootstrapped PPO and reaches parity with lightly
    bootstrapped PPO at larger batch sizes. At longer horizons and in more complex domains, GRPO
    needs larger batches to catch up to PPO's performance. Table~\ref{tab:gymnasium-crossovers} summarizes all comparisons (Section~\ref{sec:batch-critic-analysis}).

    \item \textbf{Studying the bias--variance tradeoff in terms of gradient direction helps explain the crossover.}
    We characterize each estimator by its \emph{alignment}, how closely its mean gradient points along the true policy gradient, and its \emph{noise scale}, its variance relative to the squared norm of its mean gradient.
    This analysis indicates when GRPO's batch gradient estimates point closer to the true policy gradient than PPO's as batch size grows and why lighter bootstrapping can delay this crossover (Section~\ref{sec:theory}).
    Gradient measurements in control tasks support this account and show higher noise scales at longer horizons and in
    HalfCheetah than in Pendulum (Section~\ref{sec:llm-gradient-results}).
    \item \textbf{Critics pay off most in the low-compute regime.}
    At small batches, PPO's critic can provide enough accuracy improvement to justify its additional cost. Once GRPO reaches comparable performance at larger batches, it requires less estimated training compute (Section~\ref{sec:compute-analysis}).

\end{enumerate}

\section{Preliminaries}
\label{headings}

\subsection{PPO}
\label{sec:ppo-background}
PPO estimates advantages using a learned value function or
critic~\citep{schulman2017ppo}. Consider a trajectory
collected under $\pi_{\theta_{\mathrm{old}}}$, with states $s_t$ and actions
$a_t$ for $t=1,\ldots,H$, where $H$ is the horizon. For LLMs, each action is a
generated token, and the state is the context available before generating it.
With importance-sampling ratio
$\rho_t(\theta)=\pi_\theta(a_t\mid s_t)/\pi_{\theta_{\mathrm{old}}}(a_t\mid s_t)$,
PPO maximizes the clipped surrogate objective
\begin{equation}
\mathcal{J}_{\text{PPO}}(\theta)
=\mathbb{E}_{t}\big[\min\{\rho_t\hat{A}_t,\,
\operatorname{clip}(\rho_t,1-\epsilon,1+\epsilon)\hat{A}_t\}\big],
\label{eq:ppo-objective}
\end{equation}
where $\epsilon$ sets the clipping range. The critic $V_\phi(s_t)$ estimates
the expected discounted future reward from state $s_t$, and GAE uses its
predictions to compute the advantage $\hat{A}_t$.

\paragraph{Generalized advantage estimation.}
In the finite-horizon setting, let $D=H-t+1$ denote the number of
transitions remaining and set $V_\phi(s_{H+1})=0$ at the terminal state.
For $1\le k\le D$, the $k$-step advantage estimate is
\[
\hat{A}^{(k)}_t
=-V_\phi(s_t)+\sum_{\ell=0}^{k-1}\gamma^\ell r_{t+\ell}
+\gamma^kV_\phi(s_{t+k}),
\]
where $\gamma\in[0,1]$ is the reward discount factor and $r_t$ is the reward
after action $a_t$.
GAE mixes these estimates with geometrically decaying weights controlled by
$\lambda\in[0,1]$~\citep{schulman2015gae}:
\begin{equation}
\hat{A}^{\mathrm{GAE}(\gamma,\lambda)}_t
=(1-\lambda)\sum_{k=1}^{D-1}\lambda^{k-1}\hat{A}^{(k)}_t
+\lambda^{D-1}\hat{A}^{(D)}_t.
\label{eq:gae}
\end{equation}

The estimates with $k<D$ bootstrap from the critic, while $\hat{A}^{(D)}_t$
is the Monte Carlo return minus the baseline. The final estimate receives
the remaining weight $\lambda^{D-1}$, so the weights sum to one. Smaller $\lambda$ bootstraps more,
typically reducing variance at the cost of critic-induced bias. At
$\lambda=1$, only the Monte Carlo estimate remains, with no bootstrap bias.
The critic regresses on the $\lambda_V$-return target
$\hat V_t=\hat{A}^{\mathrm{GAE}(\gamma,\lambda_V)}_t+V_\phi(s_t)$ with loss
\begin{equation}
\mathcal{L}_{\mathrm{VF}}(\phi)
=\mathbb{E}_t\big[(V_\phi(s_t)-\hat V_t)^2\big].
\label{eq:value-loss}
\end{equation}
The target $\hat V_t$ is held fixed during the critic update.
\emph{Decoupled GAE} lets the critic's $\lambda_V$ differ from the actor's
$\lambda$, typically with $\lambda_V=1$ for Monte Carlo critic
targets~\citep{yuan2025vcppo,qi2026bestpracticecriticoptimization}.

\paragraph{Length-adaptive GAE.}
With fixed $\lambda<1$, the full Monte Carlo estimate receives progressively
less weight at early steps as the trajectory grows longer, increasing
reliance on intermediate critic predictions. Following VAPO, length-adaptive
GAE adjusts $\lambda$ with the trajectory horizon $H$~\citep{yue2025vapo}:
\begin{equation}
\lambda(H)=1-\frac{1}{\alpha H},
\label{eq:length-adaptive-gae}
\end{equation}
for $\alpha H>1$. Larger $\alpha$ gives more weight to the Monte Carlo
estimate and reduces bootstrapping. At step $t$, this estimate receives
weight $\lambda(H)^{H-t}$, which is smallest at the first step. For long
trajectories, this minimum approaches $e^{-1/\alpha}$, so $\alpha$ sets an
approximately length-independent minimum Monte Carlo weight
(Appendix~\ref{app:gae-derivation}). These minimums are approximately $4.5\times10^{-5}$, $0.14$, and $0.51$
for $\alpha=0.1$, $0.5$, and $1.5$, respectively. 

\paragraph{Privileged critics.} The critic may also receive
\emph{privileged information} $z$ hidden from the policy, such as a reference
solution, and predict
$V_\phi(s_t,z)$~\citep{venkatraman2026lecritique,zhang2026redippo,qi2026bestpracticecriticoptimization}.
To keep the privileged baseline from introducing policy-gradient bias, it
suffices that $z$ be conditionally independent of the current action given
$s_t$~\citep{williams1992simple,greensmith2004variance,venkatraman2026lecritique}.

\subsection{GRPO}
\label{sec:grpo-background}
GRPO estimates advantages by comparing rewards across trajectories sampled
from the same prompt or start state~\citep{shao2024deepseekmath}. For a
group of $G>1$ trajectories collected under $\pi_{\theta_{\mathrm{old}}}$,
it maximizes Equation~\ref{eq:ppo-objective} averaged over their actions,
using advantage $\hat A_{i,t}$~\citep{yu2025dapo,liu2025drgrpo}.
With outcome reward $r_i$ for trajectory $i$ and the group's reward mean
$\bar r$ and standard deviation $\sigma_r$, GRPO centers each reward and
divides by $\sigma_r$. The division reweights groups by how much their
rewards vary, so Dr.~GRPO omits
it~\citep{liu2025drgrpo}:
\begin{equation}
\hat{A}^{\mathrm{GRPO}}_{i,t}=\frac{r_i-\bar r}{\sigma_r},
\quad
\hat{A}^{\mathrm{Dr.\,GRPO}}_{i,t}=r_i-\bar r.
\label{eq:grpo-advantage}
\end{equation}
Groups with identical rewards get zero advantage. In both forms, all actions
in a trajectory share one advantage, so GRPO provides no within-trajectory credit
assignment, and without a critic its advantages carry no critic approximation
bias.

\section{A Directional Bias--Variance Account}
\label{sec:theory}
\subsection{Bias and Variance of Gradient Estimators}
\label{sec:bias-variance-tradeoff}

Let $g=\nabla_\theta\mathbb E_{\tau\sim\pi_\theta}[r(\tau)]$ be the policy
gradient of the expected outcome reward $r(\tau)$ of a trajectory $\tau$. A batch gradient $\hat g_B$ averages $B$ per-trajectory
estimates $\hat g$, which may be biased depending on the advantage estimator.
These estimates are independent across trajectories, or across groups for
GRPO (Appendix~\ref{app:alignment}). With bias
$b=\mathbb E[\hat g]-g$ and variance $V=B\operatorname{Var}(\hat g_B)$,
the mean squared error (MSE) decomposes as
\begin{equation}
\mathbb E\!\left[\|\hat g_B-g\|^2\right]
=\underbrace{\|b\|^2}_{\text{squared bias}}
+\underbrace{V/B}_{\text{variance}}.
\label{eq:gradient-bias-variance}
\end{equation}

\paragraph{Gradient direction.} MSE measures discrepancies in both gradient direction
and magnitude, but popular optimizers such as AdamW and Muon make updates
largely invariant to gradient
magnitude~\citep{kingma2015adam,loshchilov2019adamw,liu2025muon}. We therefore focus on gradient directions independently of their
magnitude~\citep{ilyas2018closerlook}. We measure an estimator's \emph{alignment} by
the cosine similarity $c=\cos(\mathbb E[\hat g],g)$ and its 
\emph{noise scale} by $\nu=V/\|\mathbb E[\hat g]\|^2$, which
extends the simple noise scale of
\citet[Equation~2.9]{mccandlish2018largebatch} to biased estimators. These are directional counterparts of the bias and variance terms in
Equation~\ref{eq:gradient-bias-variance}. 

The cosine similarity between $\hat g_B$ and $g$ ranks unit-norm
stochastic gradient descent (SGD) steps to first order, but not updates that intentionally depart from $g$,
such as preconditioned or exploration-driven ones (Appendix~\ref{app:alignment}).

Under our shared actor objective (Section~\ref{sec:comparison-protocol}), PPO
and GRPO both estimate the same $g$, so cosine similarity to $g$ is a fair
basis for comparing them. The \emph{cosine model} approximates the
expected cosine similarity of a batch of size $B$ as
\begin{equation}
\mathbb E\!\left[\cos(\hat g_B,g)\right]\approx\frac{c}{\sqrt{1+\nu/B}}.
\label{eq:cosine-model}
\end{equation}
We derive this approximation in Appendix~\ref{app:alignment}.
When $B\ll\nu$, Equation~\ref{eq:cosine-model} gives an expected cosine
similarity of approximately $c\sqrt{B/\nu}$, so sampling noise strongly limits
the expected cosine similarity. When $B\gg\nu$, expected cosine similarity approaches $c$, and
additional sampling gives diminishing improvements.
Larger batches shrink the noise term $\nu/B$ but cannot raise the
expected cosine similarity above $c$, so $c$ acts as an alignment ceiling as
batch size grows. This can favor PPO's
lower noise scale at small batch sizes and GRPO's better-aligned mean gradient at
large ones.

Let $c_{\mathrm{PPO}}$ and $c_{\mathrm{GRPO}}$ denote the respective
alignments, and $\nu_{\mathrm{PPO}}$ and $\nu_{\mathrm{GRPO}}$
the respective noise scales.
Assume $0<c_{\mathrm{PPO}}<c_{\mathrm{GRPO}}$ and
$0<\nu_{\mathrm{PPO}}<\nu_{\mathrm{GRPO}}$. At batch sizes small relative to
both noise scales, the cosine model ranks estimators by
$c^2/\nu$. If
$c_{\mathrm{PPO}}^2/\nu_{\mathrm{PPO}}>c_{\mathrm{GRPO}}^2/\nu_{\mathrm{GRPO}}$,
PPO leads in this regime and the approximation predicts a positive crossover
batch size
\begin{equation}
B^*_{\cos}=\frac{c_{\mathrm{PPO}}^2\nu_{\mathrm{GRPO}}-c_{\mathrm{GRPO}}^2\nu_{\mathrm{PPO}}}{c_{\mathrm{GRPO}}^2-c_{\mathrm{PPO}}^2}.
\label{eq:crossover-batch}
\end{equation}

\subsection{Horizon and Gradient Variance}
\label{sec:horizon-batch-theory}

We next apply a standard Monte Carlo gradient-variance argument to
trajectories with bounded outcome rewards~\citep{papini2022smoothing}.
Horizon is one source of task complexity, so the bound motivates our
joint study of task complexity and batch size. For a trajectory $\tau$ with states and actions $s_1, a_1, s_2, \dotsc$ and horizon $H$, consider the
Monte Carlo policy-gradient estimator~\citep{williams1992simple}:
\begin{equation}
\hat g(\tau)=r(\tau)\sum_{t=1}^{H}
\nabla_\theta\log\pi_\theta(a_t\mid s_t),
\label{eq:mc-trajectory-gradient}
\end{equation}
which is unbiased, $\mathbb E[\hat g]=g$.

\begin{proposition}[Bounded outcome rewards]
\label{cor:outcome-gradient-variance}
Consider trajectories from a fixed policy and task distribution with
$r(\tau)\in[0,1]$ regardless of horizon. Suppose
$\mathbb E[\|\nabla_\theta\log\pi_\theta(a_t\mid s_t)\|^2]\le C$
for every $t$ and $\|g\|\ge m>0$, where $C$ and $m$ are independent of $H$.
Then
\begin{equation}
V\le CH=O(H),
\qquad
\nu\le\frac{CH}{\|g\|^2}-1\le\frac{CH}{m^2}-1=O(H).
\label{eq:horizon-batch-variance}
\end{equation}
\end{proposition}

Proposition~\ref{cor:outcome-gradient-variance} gives $O(H)$ bounds because
the total reward is bounded, as in our LLM verifiers and rescaled control
rewards (Appendix~\ref{app:gradient-scaling}). With accumulated rewards,
standard bounds instead grow as
$O(H^3)$~\citep{zhao2011analysis,papini2022smoothing}.
Dimensionality enters through $C$, which sums over action
dimensions~\citep{wu2018variance}.
Because batch size enters the cosine model only through $\nu/B$, a noise scale
that grows with $CH$ needs a proportionally larger batch to approach the
alignment ceiling, so the required batch size can grow with both horizon and
dimensionality.
The bound on $\nu$ assumes that $\|g\|$ does not shrink with $H$. If it did,
the bound would grow faster than linearly, but in our control tasks $\|g\|^2$
grows with $H$ (Appendix Table~\ref{tab:control-stats}).
Section~\ref{sec:llm-gradient-results} measures these quantities in control
tasks.

\subsection{From Analysis to Experiments}
\label{sec:analysis-to-experiments}
A critic can make small batches more effective by reducing the noise scale, but
bootstrapping through an imperfect critic can also lower alignment, which
larger batches cannot recover. We therefore vary batch
size and critic reliance together, testing whether additional sampling lets
GRPO catch up and whether lighter bootstrapping keeps PPO competitive. The horizon bound, together with the domain effects discussed in
Section~\ref{sec:intro}, motivates repeating this comparison across horizons
and domains. Our analysis applies most directly to our control tasks, where updates are
on-policy and unit-norm SGD makes first-order
progress depend on the update's cosine similarity with
$g$~\citep{cutkosky2020momentum}. In LLM training, off-policy updates, token averaging, and AdamW weaken this
link, although our LLM results follow the same qualitative pattern
(Section~\ref{sec:batch-critic-analysis}).

\section{Experiments}
\label{sec:setup}

We compare PPO and GRPO in control tasks, short-horizon math, and
long-horizon terminal agents, and we vary batch size, critic reliance, and task complexity jointly. For LLMs, we also assess whether critics justify their additional compute.
Table~\ref{tab:settings} outlines the experimental settings, and
Table~\ref{tab:gymnasium-crossovers} summarizes the outcomes across tested batch sizes.

\begin{table*}[!t]
\centering
\small
\setlength{\tabcolsep}{6pt}
\renewcommand{\arraystretch}{1.25}

\begin{tabular*}{0.82\textwidth}{
@{\extracolsep{\fill}}lccc@{}
}
\toprule
& Control & Math & Terminal \\
\midrule

Policy model
& MLP
& \mbox{Qwen2.5-3B Base}
& \mbox{Qwen3.5-4B} \\

Max length
& 5, 25, 200 actions
& 4{,}096 tokens
& 65{,}536 tokens \\

Max turns
& --
& 1
& 64 \\

Batch size $B$
& 4--65{,}536
& 4--4{,}096
& 128--2{,}048 \\

PPO $\alpha$
& 0.1, 0.5, 1.5
& 0.1, 0.5, 1.5
& 0.5, 1.5 \\

Gradient updates
& 5{,}000
& 800
& 200 \\

Seeds
& 5
& 3
& 1 \\

\bottomrule
\end{tabular*}

\caption{
\textbf{Experimental settings.}
Batch sizes increase by factors of four.
Max length denotes the episode cap in actions for
control tasks and in tokens for LLM tasks
(response length for math, context length for
terminal agents).
Max turns is the maximum number of assistant turns.
}
\label{tab:settings}
\end{table*}
\subsection{Shared Comparison Protocol}
\label{sec:comparison-protocol}
\label{sec:shared-objective}

\paragraph{Matched batch sizes.} Let $P$ be the number of prompts (or start states) and $G$ the number of trajectories per prompt,
so that the batch size is $B=PG$. GRPO
requires $G>1$ to form its group baseline, while PPO can use $G=1$. We compare their advantage estimators under a shared actor objective and matched trajectory batch sizes, starting from the same policy within each setting. Group sizes
differ between methods. At $B=4{,}096$, our choices of
$G=16$ for GRPO and $G=1$ for PPO are at or near the best group size by expected cosine similarity, so comparisons at the largest batch size do not hinge on this choice (Appendix~\ref{app:group-size}).

\paragraph{Crossover and parity.}
At each tested batch size $B$, one method \emph{leads} if its mean
score exceeds the other's by more than twice the standard error of
the difference. Otherwise, the methods are considered \emph{tied}. For each setting and PPO variant, we summarize the comparison across batch sizes by its outcome at the largest tested batch size:
a \emph{crossover} if GRPO leads, \emph{no crossover} if PPO leads,
and \emph{parity} if they are tied. A crossover starts at the smallest
tested batch size $B$ where GRPO leads at $B$ and every larger tested batch
size. For comparisons ending
in parity, parity starts at the smallest tested batch size $B$ where PPO
does not lead at $B$ or any larger tested batch size.
Standard errors are over training seeds in the control tasks and
math, and over evaluations for terminal agents,
where we have one training seed. Control comparisons use validation
scores, while LLM comparisons use the downstream test benchmark average.

\paragraph{Shared setup.} Following SAO, IcePop, and
INTELLECT-3~\citep{sao2026,zhao2025icepop,primeintellect2025intellect3}, all methods
maximize the masked objective
\begin{equation}
\mathcal{J}_{\mathrm{IcePop}}(\theta)=\mathbb{E}_t\!\left[
\mathcal{M}\!\left(\frac{\pi_\theta(a_t\mid s_t)}{\pi_{\mathrm{infer}}(a_t\mid s_t)};\,1-\epsilon_{\mathrm{low}},\,1+\epsilon_{\mathrm{high}}\right)\hat{A}_t\right],
\label{eq:shared-actor-objective}
\end{equation}
where $\pi_\theta$ is the policy being trained, $\pi_{\mathrm{infer}}$ is the
policy that collected the rollouts, and $\epsilon_{\mathrm{low}}=0.5$ and
$\epsilon_{\mathrm{high}}=4.0$ set the range of probability ratios that
$\mathcal{M}$ keeps. In control tasks,
updates are on-policy with no training--inference
mismatch, so masking is inactive.
PPO uses decoupled GAE with length-adaptive $\lambda$ to compute $\hat A_t$, normalizes advantages over each batch, and uses $\lambda_V=1$ for critic updates (Section~\ref{sec:ppo-background}).
Our critics also see privileged inputs hidden from the actor, and
Appendix~\ref{app:priv-ablation} assesses their effect in math. We evaluate both Dr.~GRPO and GRPO with standard-deviation normalization in control tasks and only GRPO with standard-deviation normalization in
the LLM settings (Equation~\ref{eq:grpo-advantage}).

\subsection{Experimental Settings}
\label{sec:experimental-settings}

\begin{table*}[!t]
\centering
\footnotesize
\setlength{\tabcolsep}{4pt}
\renewcommand{\arraystretch}{1.25}

\begin{tabularx}{0.82\textwidth}{
@{}lc*{3}{>{\centering\arraybackslash}X}
}
\toprule
& & \multicolumn{3}{c}{PPO ($\alpha$)} \\
\cmidrule(lr){3-5}
Setting & $H$ & $0.1$ & $0.5$ & $1.5$ \\
\midrule

\multirow{3}{*}{Pendulum}
& $5$
& \cellcolor[HTML]{F5BB8C}GRPO ($4$)
& \cellcolor[HTML]{F5BB8C}GRPO ($4$)
& \cellcolor[HTML]{F5BB8C}GRPO ($4$) \\

& $25$
& \cellcolor[HTML]{F9D6B8}GRPO ($256$)
& \cellcolor[HTML]{FBE1CC}GRPO ($1\text{k}$)
& \cellcolor[HTML]{FBE1CC}GRPO ($1\text{k}$) \\

& $200$
& \cellcolor[HTML]{FFF5EC}GRPO ($65\text{k}$)
& \cellcolor[HTML]{F2F2F2}Parity ($65\text{k}$)
& \cellcolor[HTML]{F2F2F2}Parity ($65\text{k}$) \\

\addlinespace[5pt]

\multirow{3}{*}{HalfCheetah}
& $5$
& \cellcolor[HTML]{F7C49C}GRPO ($16$)
& \cellcolor[HTML]{F7C49C}GRPO ($16$)
& \cellcolor[HTML]{F7C49C}GRPO ($16$) \\

& $25$
& \cellcolor[HTML]{FCEADB}GRPO ($4\text{k}$)
& \cellcolor[HTML]{F2F2F2}Parity ($16\text{k}$)
& \cellcolor[HTML]{E6EEF7}PPO \\

& $200$
& \cellcolor[HTML]{F2F2F2}Parity ($256$)
& \cellcolor[HTML]{E6EEF7}PPO
& \cellcolor[HTML]{E6EEF7}PPO \\

\midrule

Math
& --
& \cellcolor[HTML]{FBE1CC}GRPO ($1\text{k}$)
& \cellcolor[HTML]{F2F2F2}Parity ($1\text{k}$)
& \cellcolor[HTML]{F2F2F2}Parity ($256$) \\

\midrule

Terminal
& --
& --
& \cellcolor[HTML]{F2F2F2}Parity ($2\text{k}$)
& \cellcolor[HTML]{F2F2F2}Parity ($2\text{k}$) \\

\bottomrule
\end{tabularx}

\vspace{6pt}
{\footnotesize\setlength{\fboxsep}{0pt}%
\newcommand{\legendSwatch}[1]{%
  \colorbox[HTML]{#1}{\rule{0pt}{1.3ex}\hspace{1.25em}}%
}%
GRPO crossover, small $B$\
\legendSwatch{F5BB8C}%
\legendSwatch{F7C49C}%
\legendSwatch{F9D6B8}%
\legendSwatch{FBE1CC}%
\legendSwatch{FCEADB}%
\legendSwatch{FFF5EC}\ large $B$%
\qquad\legendSwatch{F2F2F2}\ Parity%
\qquad\legendSwatch{E6EEF7}\ PPO leads\par}

\caption{
\textbf{PPO--GRPO outcomes across batch sizes.}
\emph{GRPO ($B$)} denotes a crossover: GRPO leads
at $B$ and every larger tested batch.
\emph{Parity ($B$)} indicates that PPO no longer leads
from $B$ onward, with a tie at the largest tested batch.
\emph{PPO} indicates a continued PPO lead at the
largest tested batch.
Darker orange marks earlier GRPO crossovers;
gray denotes parity and blue a continued PPO lead.
Control results use Dr.~GRPO validation scores,
and LLM results use downstream test averages.
See Section~\ref{sec:comparison-protocol} for the
statistical criterion. A dash indicates an untested
variant or inapplicable horizon.
}
\label{tab:gymnasium-crossovers}

\end{table*}

\paragraph{Control tasks.}
\label{sec:controlled-continuous-control}
We use fixed-horizon versions of Gymnasium's Pendulum-v1 and
HalfCheetah-v4~\citep{towers2024gymnasium}, with horizons $H\in\{5,25,200\}$ and
a two-layer multilayer perceptron (MLP) actor. We accumulate the environment's
per-step rewards over each episode, rescale their average to $[0,1]$, and
provide the reward only at the final state.
GRPO uses $G=16$ trajectories per start state, except at $B=4$, where $G=4$.
Both methods take one unit-norm SGD update per batch.
Actor and critic learning rates are $10^{-3}$ and $5\times10^{-3}$,
respectively. We focus on Dr.~GRPO because
standard-deviation normalization can overweight uninformative groups, especially with continuous
rewards, where group reward standard deviations can differ by orders of magnitude. Appendix~\ref{app:std-normalization} analyzes this effect and reports results for GRPO with standard-deviation normalization and PPO with GAE $\lambda = 1$. We report each run's best validation score. Appendix~\ref{app:control-networks} gives the remaining environment and training details.

\begin{figure*}[t!]
\centering
\includegraphics{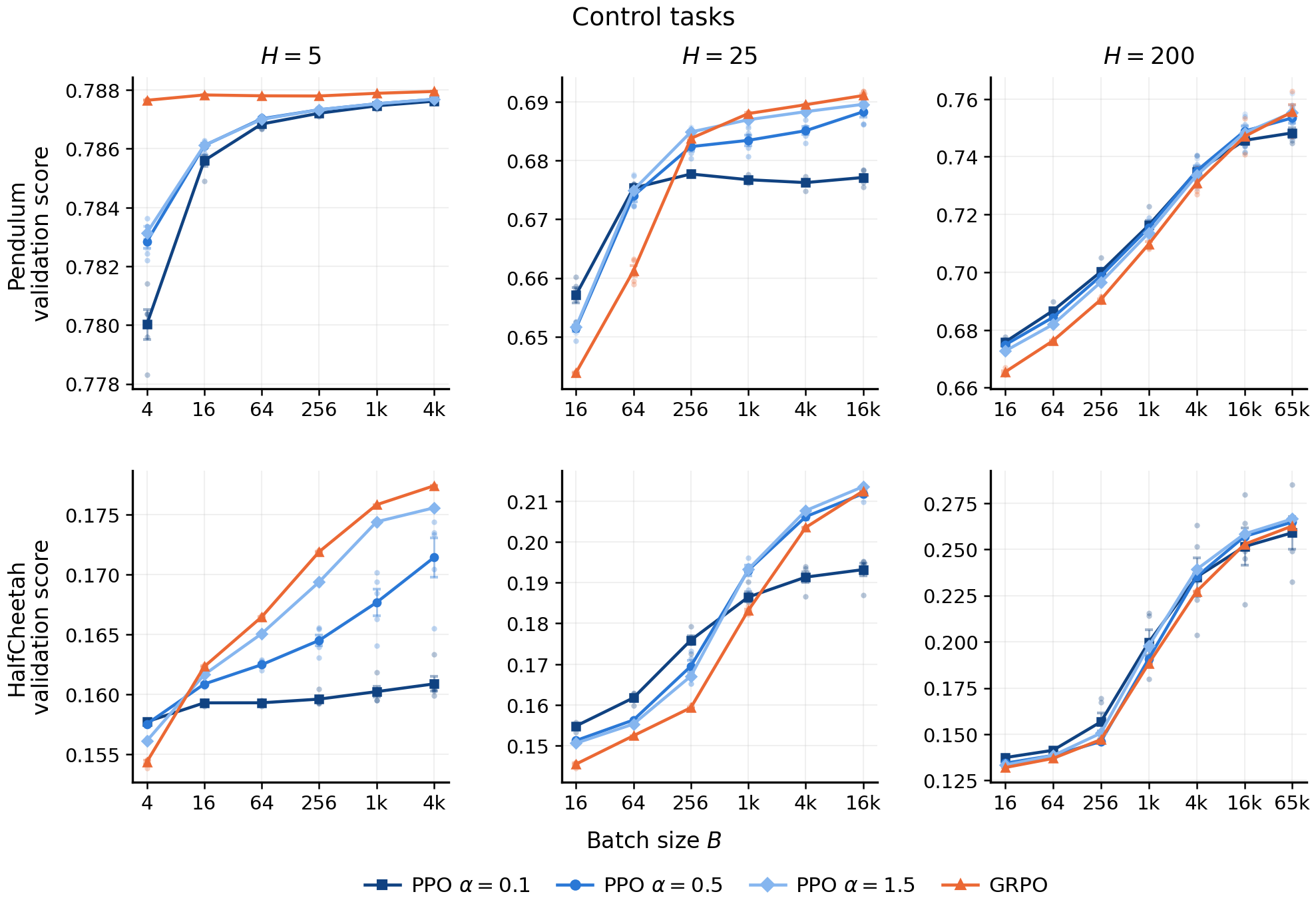}
\caption{\textbf{GRPO gains on PPO as batches grow.}
Best validation score within 5,000 updates in Pendulum (top) and HalfCheetah
(bottom). GRPO uses the Dr.~GRPO advantage, without standard-deviation normalization. Mean and one standard error over
five seeds, with dots for individual seeds.}
\label{fig:gymnasium-batch-scaling}
\end{figure*}

\begin{figure*}[t!]
\centering
\includegraphics{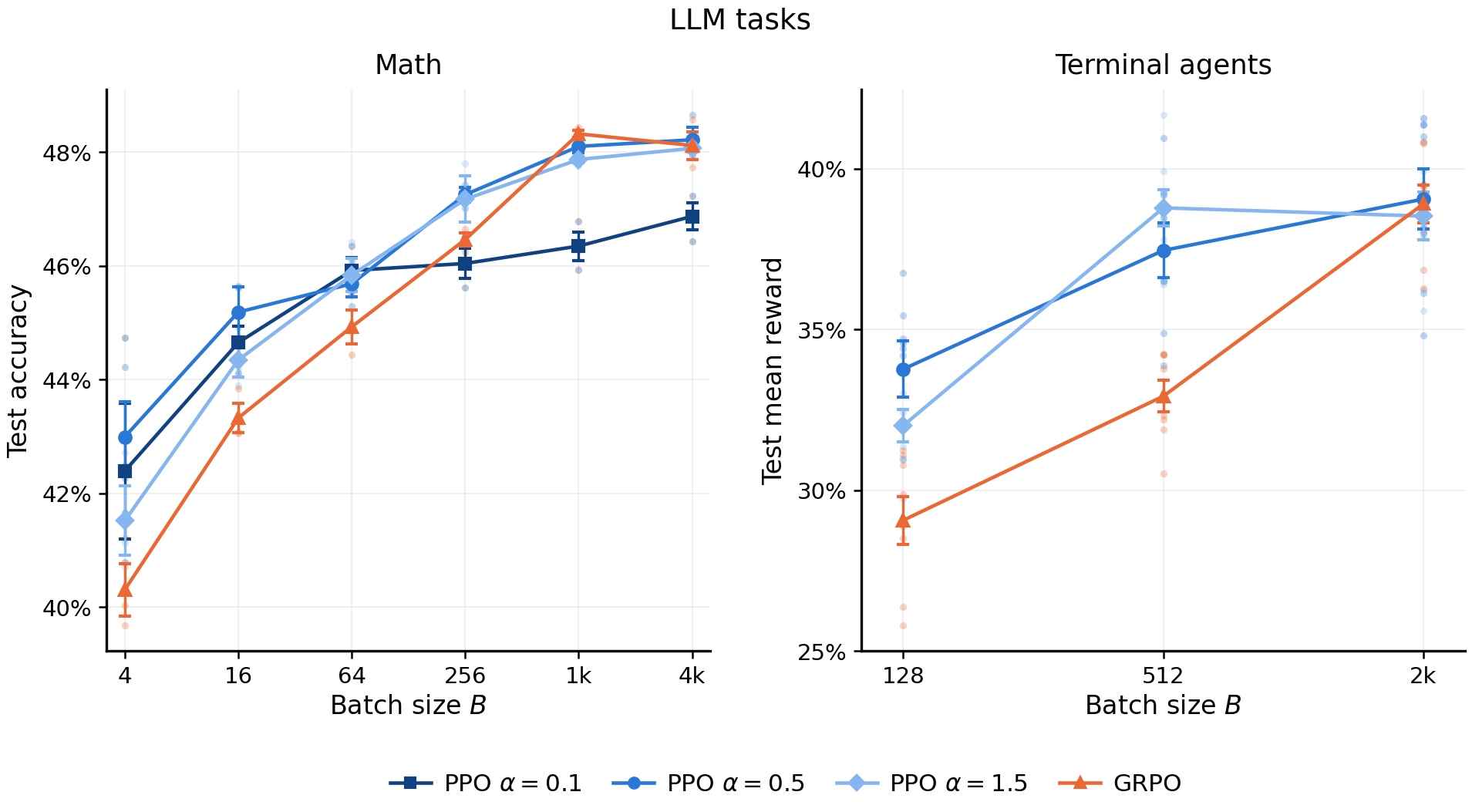}
\caption{\textbf{PPO's lead fades as batches grow in LLM tasks.} Test
benchmark average with mean and one standard error. Left: five math benchmarks
at each run's best validation checkpoint, over three training seeds. Right:
Terminal-Bench 2.1 and TB-Lite at the final checkpoint, with one training seed
and eight evaluation passes. Faint dots show individual seeds or passes, and
Appendix Tables~\ref{tab:math-all-scores} and~\ref{tab:terminal-all-scores} list validation scores.}
\label{fig:short-math-best}
\label{fig:terminal-validation}
\end{figure*}

\paragraph{Short-horizon math.}
\label{sec:llm-tasks}
\label{sec:short-math}
We train Qwen2.5-3B Base on 10,000 DeepScaleR
problems~\citep{qwen2025qwen25,deepscaler2025}, with a binary reward for
final-answer correctness. GRPO uses $G=\min(16,\sqrt{B})$ trajectories per
problem. Each step samples $2B$ trajectories for two actor updates of batch
size $B$, and PPO takes one critic update per actor update. Actor and
critic learning rates are $10^{-6}$ and $2.5\times10^{-6}$, respectively, with
AdamW~\citep{loshchilov2019adamw}. PPO's critic is pretrained and receives
the reference answer and a solution as privileged inputs.
We select each run's best checkpoint
by avg@16 on 500 held-out
DeepScaleR problems and evaluate it on MATH-500, GSM8K, AMC23, Minerva
Math, and OlympiadBench~\citep{hendrycks2021math,lightman2024verify,cobbe2021gsm8k,lewkowycz2022minerva,he2024olympiadbench},
reporting mean accuracy over multiple samples per problem.
Appendix~\ref{app:llm-data} provides additional training and evaluation details.

\paragraph{Long-horizon terminal agents.}
\label{sec:terminal-agents}
We train Qwen3.5-4B on TMax-15K terminal
tasks~\citep{qwen2026qwen35,ivison2026tmax}, largely following the Tmax setup. The agent runs bash commands in
a sandbox, and hidden tests on the final environment state assign the reward.
GRPO uses $G=16$ trajectories per task.
Training is asynchronous, with one actor update per step and a maximum
staleness of four policy versions. Actor and critic learning rates are
$10^{-6}$ and $5\times10^{-6}$, respectively, with AdamW. PPO's critic is
pretrained for 250 updates of 64 trajectories, and its privileged inputs are the task's grading tests and
target state. We report all metrics
at the final checkpoint (step 200): validation avg@3 on 500 held-out
TMax-15K tasks and test avg@8 on Terminal-Bench 2.1 and
TB-Lite~\citep{merrill2026terminalbench,openthoughts2026tblite}.
Appendix~\ref{app:llm-data} gives the remaining LLM training and evaluation
details, including the differences from Tmax.

\subsection{Batch Size, Critic Reliance, and Task Complexity}
\label{sec:batch-critic-analysis}
\label{sec:cross-setting-analysis}
\label{sec:gymnasium-results}
\label{sec:short-math-results}
\label{sec:terminal-results}
\label{sec:complexity-analysis}

\paragraph{PPO leads at small batch sizes, and GRPO catches up as batches grow.}
Figure~\ref{fig:gymnasium-batch-scaling} shows that PPO leads at the smallest
batch sizes in every control setting except Pendulum at $H=5$, where GRPO
leads throughout. As batch size grows, GRPO closes the gap, and at the largest
tested batch sizes it leads or ties PPO in 15 of 18 control comparisons
(Table~\ref{tab:gymnasium-crossovers}). The LLM
settings follow the same pattern (Figure~\ref{fig:short-math-best}). In math,
the best PPO variant leads GRPO at the two smallest batch sizes,
$B\in\{4,16\}$, by 1.9--2.7 points, and GRPO leads or ties every PPO variant
at the two largest, $B\in\{1{,}024, 4{,}096\}$. In terminal agents, both PPO
variants lead GRPO at the two smaller batch sizes, $B\in\{128, 512\}$, by up
to 5.9 points, and all three methods tie at the largest, $B=2{,}048$, within
0.6 points (Appendix Tables~\ref{tab:math-all-scores}
and~\ref{tab:terminal-all-scores}). A critic
therefore helps most when samples are scarce.

\paragraph{Lighter bootstrapping keeps PPO competitive at large batch sizes.}
Whether GRPO crosses over at large batch sizes depends on how heavily PPO
bootstraps. Against heavily
bootstrapped PPO ($\alpha=0.1$), GRPO crosses over in six of seven settings,
and the exception, HalfCheetah at $H=200$, ends in parity. In math, GRPO leads
from $B=1{,}024$ onward. In contrast, against lightly bootstrapped PPO
($\alpha\in\{0.5,1.5\}$), GRPO
crosses over only in Pendulum at $H\in\{5,25\}$ and HalfCheetah at $H=5$, and
every other comparison ends in parity or a continued PPO lead. This pattern is
what Section~\ref{sec:theory} predicts. Larger batches shrink the
noise term but leave alignment unchanged, so heavy bootstrapping loses its advantage first.
Removing bootstrapping entirely changes little. In control tasks, PPO with
$\lambda=1$, which uses the critic only as a baseline, scores within $0.004$ of
PPO with $\alpha=1.5$ at every batch size (Appendix~\ref{app:all-control}).

\paragraph{Longer horizons and more complex domains delay the crossover.}
In Pendulum with $\alpha=0.1$, the crossover moves from $B=4$ at $H=5$ to $256$
at $H=25$ and $65{,}536$ at $H=200$ (Table~\ref{tab:gymnasium-crossovers}).
With lighter bootstrapping, Pendulum's longest horizon ends in parity, and
HalfCheetah's crossovers disappear beyond
$H=5$. Changing the domain has the same effect. At every matched horizon and
$\alpha$, HalfCheetah's outcome is less favorable to GRPO than Pendulum's.
Against the same two lightly bootstrapped PPO variants, terminal agents reach
parity at $B=2{,}048$, whereas math reaches it at $1{,}024$ and $256$, respectively. The
noise scale shows the same ordering, rising with horizon and $4$--$11\times$
larger in HalfCheetah than in Pendulum (Section~\ref{sec:llm-gradient-results}).
Because the domains also differ in dynamics, rewards, and training,
these comparisons suggest a role for task complexity without isolating it.

\begin{figure*}[t]
\centering
\includegraphics{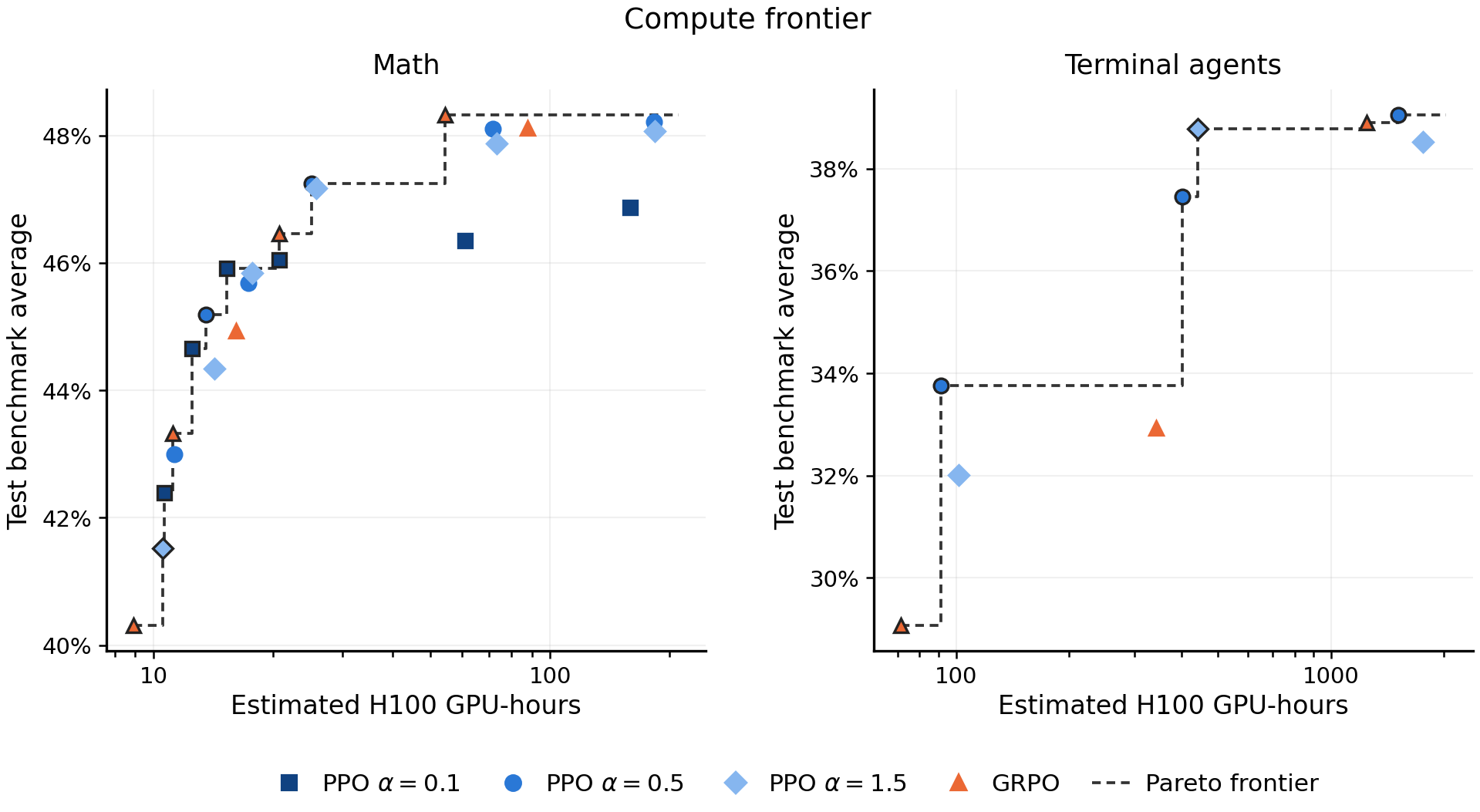}
\caption{\textbf{Critics pay off only at small compute budgets.} Test benchmark average against estimated
H100 GPU-hours for every method and batch size. Left: math over
400 steps, averaged over three training seeds. Right: terminal agents over 200
steps, with one training seed. Costs count only the GPU time that each part of
the training system needs, which for terminal agents accounts for sandbox waits
between turns. Dashed lines are the empirical Pareto frontiers of the plotted
means (Appendix~\ref{app:llm-data}).}
\label{fig:short-math-compute}
\label{fig:terminal-compute}
\end{figure*}

\subsection{Compute Efficiency in LLMs}
\label{sec:compute-analysis}

\paragraph{Critics pay off only at small compute budgets.}
Batch size also sets the cost of training. Figure~\ref{fig:short-math-compute} plots the test benchmark average against estimated H100 GPU-hours, counting only the GPU
time that each part of the training system needs, which for terminal agents
accounts for sandbox waits between turns (Appendix~\ref{app:llm-data}). The
sandboxes' own CPU time is not counted, but it would add to each terminal run
roughly in proportion to its batch size, since all terminal runs average 26 to
30 turns per trajectory. All runs use the update budgets in Table~\ref{tab:settings}. At small budgets, PPO reaches a given
accuracy with less compute. In terminal agents, PPO with $\alpha=0.5$ at
$B=128$ matches GRPO at $B=512$ with about a quarter of the compute, 91 versus
342 estimated GPU-hours, and in math, PPO with $\alpha=0.1$ lies on the Pareto
frontier at small batch sizes. At large budgets, the ordering reverses. In
math, GRPO at $B=1{,}024$ reaches the highest test mean with less compute than
any larger-batch run, and in terminal agents at $B=2{,}048$, GRPO matches both
PPO variants with 18--29\% less compute (Appendix~\ref{app:llm-data}). When
samples are scarce, the sampling a critic saves outweighs its cost, but once
GRPO reaches parity at large batch sizes, the critic's cost buys little and
can sometimes hurt.

Together, these results favor choosing batch size and critic reliance jointly
for the task and budget at hand. Horizon and domain set where PPO keeps an
advantage, and compute decides whether that advantage is worth paying for.
Section~\ref{sec:llm-gradient-results} measures the alignment and noise scale behind these outcomes.

\section{Gradient Measurements in Control Tasks}
\label{sec:llm-gradient-results}

\begin{figure*}[t]
\centering
\includegraphics{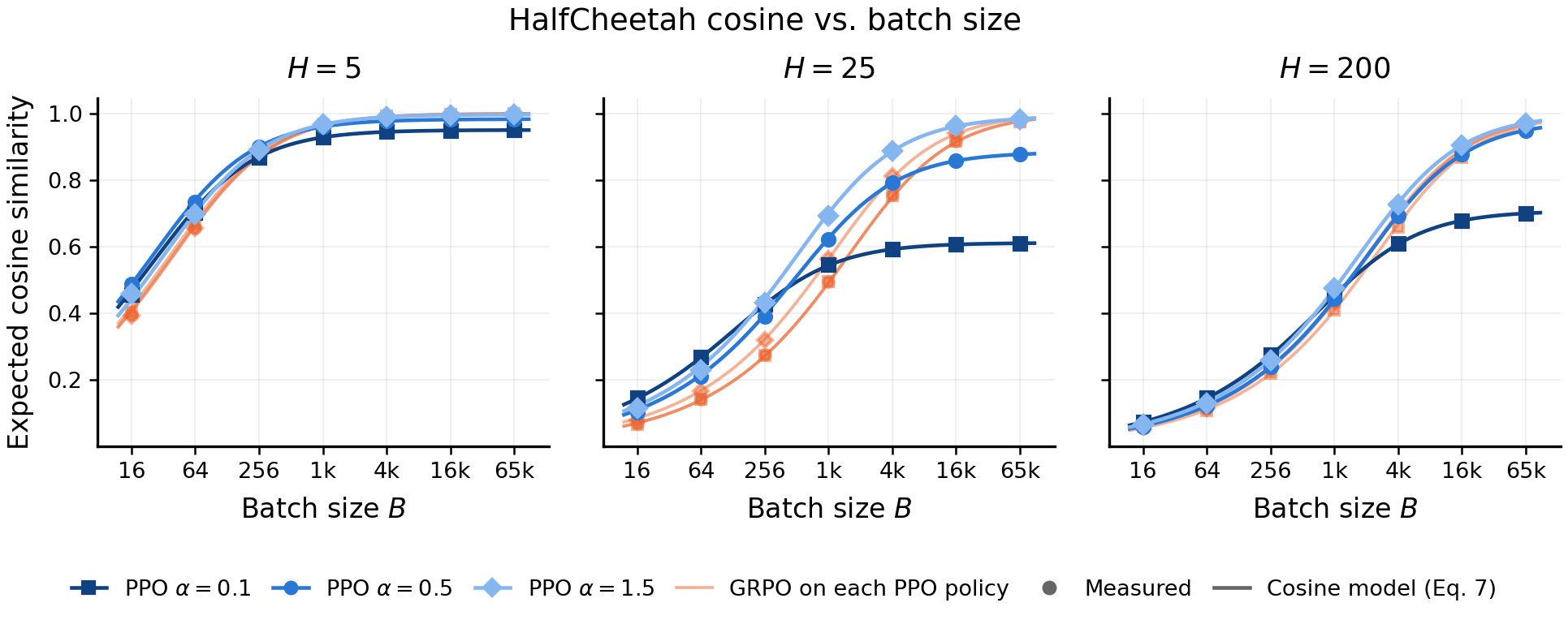}
\caption{\textbf{The cosine model tracks measured cosines.}
Expected cosine to the true policy gradient against batch size in HalfCheetah
at update 1,000, before any run's score plateaus. Markers are measurements and lines are
Equation~\ref{eq:cosine-model} with $c$ and $\nu$ estimated at $B=65{,}536$.
Each PPO variant is evaluated on its own run trained at $B=1{,}024$. Faint
orange lines evaluate GRPO on each of these three policies, with markers
matching the paired PPO variant. Appendix Figure~\ref{fig:eq7-fit} shows all
checkpoints in both environments.}
\label{fig:control-cosine-batch}
\end{figure*}

Training results show where GRPO catches up to PPO as batch size grows, but not why.
Section~\ref{sec:theory} attributes the outcome to each estimator's alignment
$c$ and noise scale $\nu$, and this section measures both in control tasks,
where an unbiased estimate from about two million trajectories gives a
reference policy gradient. The measurements confirm three parts of the account in Section~\ref{sec:theory}: alignment and noise scale determine cosine similarity
at every batch size, critics trade alignment for a lower noise scale, and the
noise scale grows with task complexity. We focus on
HalfCheetah and bring in Pendulum to compare domains. At fixed checkpoints of
each training run, we evaluate PPO's and GRPO's estimators on the same policy,
which separates gradient estimation from differences between runs
(Appendix~\ref{app:control-gradient}).

\paragraph{Alignment and noise scale predict cosine similarity at every batch size.}
We estimate $c$ and $\nu$ once, from batches of $65{,}536$ trajectories, and
use Equation~\ref{eq:cosine-model} to predict the expected cosine at six
smaller batch sizes. Figure~\ref{fig:control-cosine-batch} shows that the
predicted curves pass through the measurements for every estimator and horizon
in HalfCheetah. Across all checkpoints and estimators, the median absolute
error is $0.003$ in HalfCheetah and $0.015$ in Pendulum (Appendix
Figure~\ref{fig:eq7-fit}). An estimator's $c$ and $\nu$ are therefore enough to describe how its cosine
similarity scales with batch size.

\begin{figure*}[t]
\centering
\includegraphics{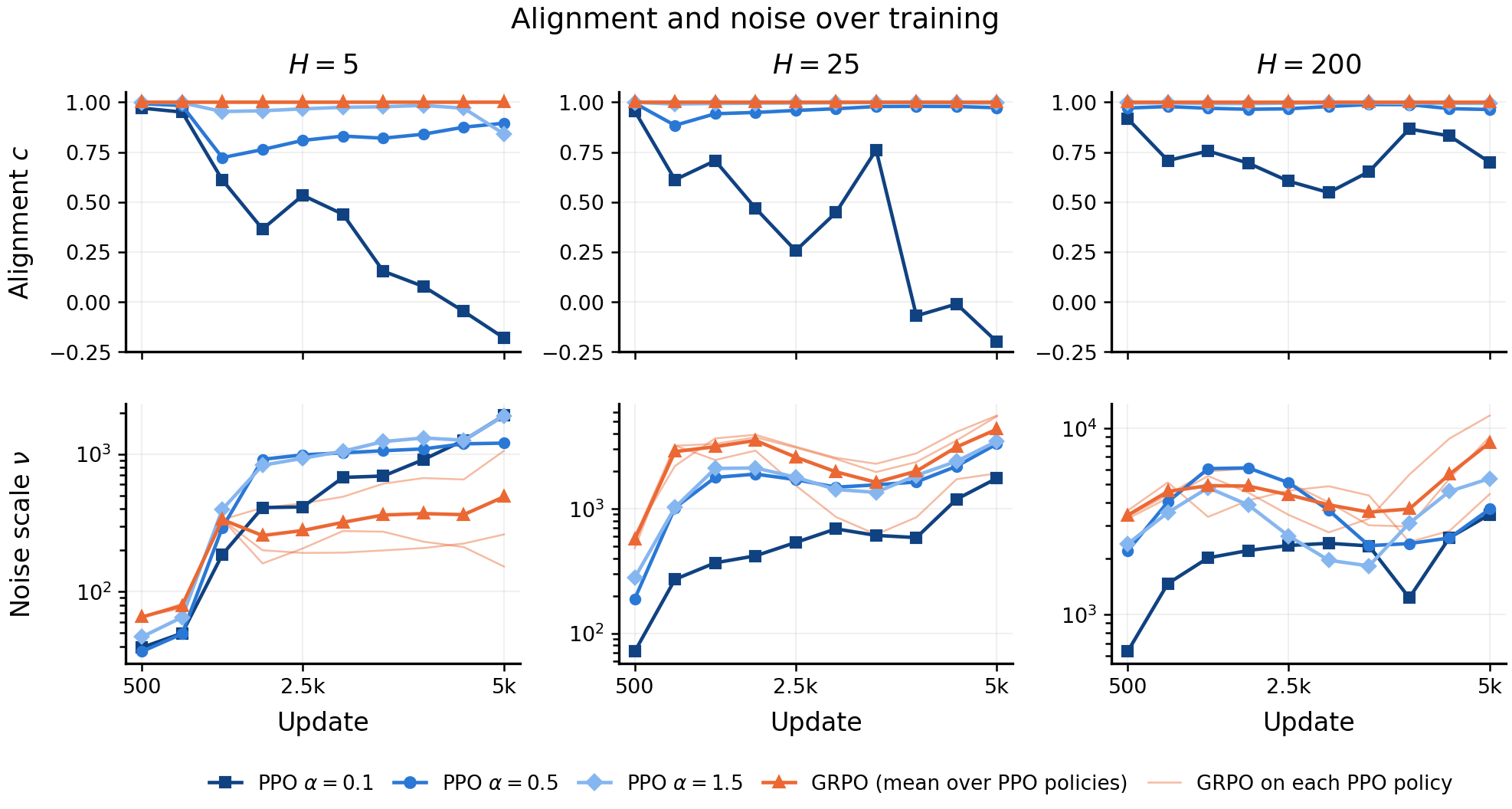}
\caption{\textbf{Heavy bootstrapping loses alignment during training.}
Alignment $c$ (top) and noise scale $\nu$ (bottom, log scale) in HalfCheetah.
Each PPO variant uses one seed trained at $B=1{,}024$, measured every 500
updates. Thin lines evaluate GRPO's estimator on each PPO policy, and the
thick line is their mean.}
\label{fig:control-training-bv}
\end{figure*}

\paragraph{In HalfCheetah, PPO trades alignment for a lower noise scale.}
Equation~\ref{eq:crossover-batch} predicts a crossover only when PPO pairs lower
alignment with a lower noise scale, and Figure~\ref{fig:control-training-bv}
shows this tradeoff in HalfCheetah. Across updates 500 to 5,000, every PPO estimator has lower alignment than GRPO's, and PPO has the lower noise scale in 66 of 90 checkpoint comparisons. Heavily bootstrapped PPO loses more alignment as training proceeds, while lighter bootstrapping gives up less,
so the crossover moves to larger batch sizes. These
fixed-policy measurements explain the direction of the tradeoff.

\begin{figure*}[t]
\centering
\includegraphics{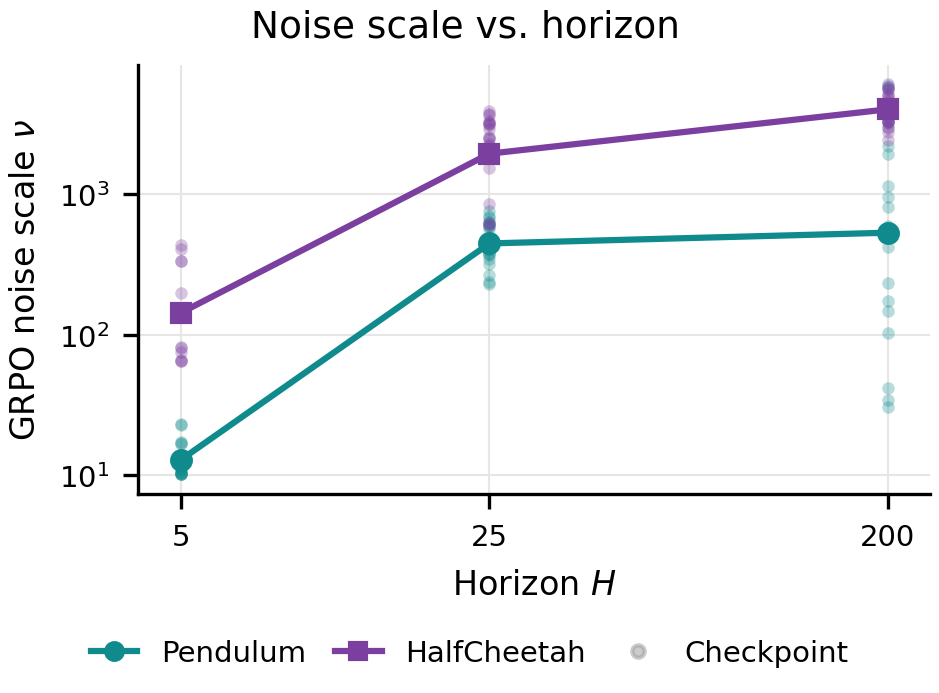}
\caption{\textbf{Noise scale grows with task complexity.}
GRPO's noise scale $\nu$ against horizon, evaluated on the policies of the
three PPO runs, one seed each. Points are checkpoints from update 500 until
each run's score plateau, and lines join geometric means. Pendulum uses
training batch sizes $16$, $64$, and $1{,}024$ at $H=5$, $25$, and $200$, and
HalfCheetah uses $1{,}024$.}
\label{fig:control-noise-horizon}
\end{figure*}

\paragraph{Noise scale grows with task complexity, in both horizon and domain.}
Figure~\ref{fig:control-noise-horizon} shows that GRPO's noise scale rises with
horizon in both environments, from $144$ to $4{,}032$ in HalfCheetah and from
$13$ to $533$ in Pendulum (Appendix~\ref{app:control-checks} explains the choice
of estimator and checkpoints). The more complex domain also has the larger noise
scale. At every horizon, HalfCheetah's is $4$--$11\times$ Pendulum's, and its
per-step score-function bound $C$ is $2.5$--$8\times$ larger, which raises the variance bound of Proposition~\ref{cor:outcome-gradient-variance}. Intuitively, larger state and action dimensionality and longer horizons mean more diverse trajectories, so a batch needs more samples to cover the relevant ones. The noise scale measures this: GRPO needs
a larger batch before its noise term $\nu/B$ becomes small, consistent with the later crossovers at longer horizons
and in HalfCheetah. Likewise, multi-turn terminal agents, whose trajectories
branch at many more intermediate decisions, reach parity at a larger batch than
short-horizon, single-turn math (Section~\ref{sec:batch-critic-analysis}).

\section{Related Work}
\label{gen_inst}

\paragraph{Critic-based versus critic-free comparisons.}
Prior comparisons find PPO stronger than GRPO in longer-horizon control and
multi-turn agents~\citep{deoliveira2025learningwithoutcritics,li2025turnppo},
with rankings also varying across tasks~\citep{wang2025ragen}.
Group Policy Gradient finds that increasing parallel rollout counts can
favor sampled baselines over PPO in control tasks~\citep{chen2025gpg}.
EVPO selects learned or sampled baselines using explained
variance~\citep{evpo2026}, while Tether mixes
them~\citep{venkatraman2026lecritique}. Other critic-based methods reduce the
critic's bootstrapping bias or cost. SAE bootstraps only at segment
boundaries~\citep{gong2026sae}, SPPO uses a sequence-level value baseline with
one sample per prompt~\citep{wang2026sppo}, and SAPO shares the policy's
backbone with the critic~\citep{liang2026sapo}. Critic-free methods can
instead obtain denser credit from implicit process rewards or
self-reflection~\citep{cui2025prime,liu2026srpo}. We examine how the effect of batch size depends on critic reliance and task complexity, and test it in LLM
training through matched comparisons in math and
multi-turn terminal agents.

\paragraph{Gradient estimation and batch scaling.}
Policy-gradient studies examine gradient variance and
direction~\citep{ilyas2018closerlook}, and gradient noise scale
characterizes diminishing batch-size returns~\citep{mccandlish2018largebatch}.
We use these quantities to explain when critic-based and critic-free estimators cross over, and compare this account with training outcomes. Batch-invariance studies show that learning-rate and
policy-update adjustments can preserve sample efficiency over a bounded
batch-size range~\citep{hilton2022batchinvariance,li2026largerbatches}.
Prior LLM batch-scaling experiments study PPO and GRPO, each on a different
model~\citep{li2026largerbatches}. IsoCompute studies
budget-dependent rollout allocation~\citep{cheng2026isocompute}, and ScaleRL
finds that larger batches train more slowly at first but reach higher
asymptotic performance~\citep{khatri2025artofscaling}.
These studies show that batch-scaling behavior depends on the training
protocol. We compare PPO and GRPO under a shared actor objective to examine how batch size and critic reliance change their relative performance.
Methods that change this objective, for example by reshaping importance
weights~\citep{chen2026respo}, are complementary.

\section{Discussion and Conclusion}
\label{sec:conclusion}

\paragraph{Longer tasks still end in a single reward.}
LLM tasks have grown from single-turn math answers to agentic trajectories
spanning dozens of turns and tens of thousands of tokens, yet under outcome
rewards each trajectory still ends in one scalar, often a pass or fail. Our
results show the cost of this mismatch. Spreading one reward across more
actions raises the noise scale
(Proposition~\ref{cor:outcome-gradient-variance} and
Section~\ref{sec:llm-gradient-results}), and the batch size at which GRPO
catches up grows with horizon and domain complexity
(Section~\ref{sec:batch-critic-analysis}). A critic is one way to extract more
signal from each trajectory, by assigning credit to individual steps rather
than spreading one reward across all of them.

\paragraph{Costly rollouts favor learning more from each one.}
Generation already dominates RL compute. In our terminal-agent runs, it
accounts for about 80\% of GRPO's estimated GPU-hours, and agentic rollouts add
costs beyond the GPU, such as sandboxes, tool and API calls, and LLM judges.
When rollouts are expensive and updates comparatively cheap, spending more
training compute to learn more from each trajectory becomes attractive.
Critics are a natural candidate. Even counting the critic's own training, they
already pay off at small compute budgets (Section~\ref{sec:compute-analysis}),
while at large budgets GRPO matches PPO with less compute, so the balance
depends on how costly rollouts are relative to updates.

\paragraph{Limitations.}
We use one starting model within each LLM setting and do not study how actor
and critic size affect crossover points. The two LLM settings also differ in several sources of task complexity at once, so unlike our control experiments they do
not isolate the effect of horizon. All runs use fixed step budgets. Our
analysis concerns gradient estimates at a fixed policy and critic. It does not
guarantee performance over multiple updates, during which PPO and GRPO may
follow different training paths. Alignment also does not rank methods that
intentionally depart from $g$, such as natural gradients or entropy
regularization. Our gradient measurements test the directional bias--variance
account only in control tasks.

\paragraph{Conclusion.}
Neither PPO nor GRPO is uniformly better across the settings we study. Our
directional bias--variance account suggests why larger batch sizes can favor
GRPO's gradient estimates. Training experiments show PPO generally leading at
small batch sizes, with GRPO often overtaking heavily bootstrapped PPO and reaching parity with lightly bootstrapped PPO at larger ones. Longer
horizons and more complex domains push these changes to larger batch sizes. We
recommend that practitioners treat batch size and critic reliance as a joint
choice based on their target task and compute budget, comparing both
performance and compute cost. As trajectories grow longer and more expensive to
collect, this choice becomes increasingly consequential.

\bibliography{references}

@article{schulman2017ppo,
  title={Proximal Policy Optimization Algorithms},
  author={Schulman, John and Wolski, Filip and Dhariwal, Prafulla and Radford, Alec and Klimov, Oleg},
  journal={arXiv preprint arXiv:1707.06347},
  year={2017}
}

@inproceedings{schulman2015gae,
  title={High-Dimensional Continuous Control Using Generalized Advantage Estimation},
  author={Schulman, John and Moritz, Philipp and Levine, Sergey and Jordan, Michael and Abbeel, Pieter},
  booktitle={International Conference on Learning Representations},
  year={2016}
}

@inproceedings{ouyang2022instructgpt,
  title={Training Language Models to Follow Instructions with Human Feedback},
  author={Ouyang, Long and Wu, Jeffrey and Jiang, Xu and others},
  year={2022},
  booktitle={Advances in Neural Information Processing Systems},
  volume={35},
  pages={27730--27744}
}

@article{shao2024deepseekmath,
  title={{DeepSeekMath}: Pushing the Limits of Mathematical Reasoning in Open Language Models},
  author={Shao, Zhihong and Wang, Peiyi and Zhu, Qihao and Xu, Runxin and Song, Junxiao and Bi, Xiao and Zhang, Haowei and Zhang, Mingchuan and Li, Y. K. and Wu, Y. and Guo, Daya},
  journal={arXiv preprint arXiv:2402.03300},
  year={2024}
}

@article{guo2025deepseekr1,
  title={{DeepSeek-R1} Incentivizes Reasoning in {LLMs} through Reinforcement Learning},
  author={Guo, Daya and Yang, Dejian and Zhang, Haowei and others},
  journal={Nature},
  volume={645},
  number={8081},
  pages={633--638},
  year={2025},
  doi={10.1038/s41586-025-09422-z}
}

@inproceedings{towers2024gymnasium,
  title={{Gymnasium}: A Standard Interface for Reinforcement Learning Environments},
  author={Towers, Mark and Kwiatkowski, Ariel and Balis, John U. and De Cola, Gianluca and Deleu, Tristan and Goul{\~a}o, Manuel and others},
  booktitle={Advances in Neural Information Processing Systems, Datasets and Benchmarks Track},
  year={2025},
  volume={38}
}

@article{ivison2026tmax,
  title={{Tmax}: A Simple Recipe for Terminal Agents},
  author={Ivison, Hamish and Yin, Junjie Oscar and Shao, Rulin and Xiao, Teng and Lambert, Nathan and Hajishirzi, Hannaneh},
  journal={arXiv preprint arXiv:2606.23321v1},
  year={2026}
}

@inproceedings{hu2025openreasonerzero,
  title={{Open-Reasoner-Zero}: An Open Source Approach to Scaling Up Reinforcement Learning on the Base Model},
  author={Hu, Jingcheng and Zhang, Yinmin and Han, Qi and Jiang, Daxin and Zhang, Xiangyu and Shum, Heung-Yeung},
  booktitle={Advances in Neural Information Processing Systems},
  volume={38},
  year={2025}
}

@article{yuan2025vcppo,
  title={What's Behind {PPO}'s Collapse in {Long-CoT}? {V}alue Optimization Holds the Secret},
  author={Yuan, Yufeng and Yue, Yu and Zhu, Ruofei and Fan, Tiantian and Yan, Lin},
  journal={arXiv preprint arXiv:2503.01491},
  year={2025}
}

@article{yue2025vapo,
  title={{VAPO}: Efficient and Reliable Reinforcement Learning for Advanced Reasoning Tasks},
  author={Yue, Yu and Yuan, Yufeng and Yu, Qiying and others},
  journal={arXiv preprint arXiv:2504.05118},
  year={2025}
}

@inproceedings{liu2025asyppo,
  title={Asymmetric Proximal Policy Optimization: Mini-Critics Boost {LLM} Reasoning},
  author={Liu, Jiashun and Obando-Ceron, Johan S. and Lu, Han and He, Yancheng and Wang, Weixun and Su, Wenbo and Zheng, Bo and Castro, Pablo Samuel and Courville, Aaron and Pan, Ling},
  booktitle={International Conference on Learning Representations},
  year={2026}
}

@article{gong2026sae,
  title={Segmental Advantage Estimation: Enhancing {PPO} for Long-Context {LLM} Training},
  author={Gong, Xue and Yi, Qi and Nan, Ziyuan and Huang, Guanhua and Li, Kejiao and Jiang, Yuhao and Xiong, Ruibin and Xu, Zenan and Guo, Jiaming and Peng, Shaohui and Zhou, Bo},
  journal={arXiv preprint arXiv:2601.07320},
  year={2026}
}

@inproceedings{kazemnejad2024vineppo,
  title={{VinePPO}: Refining Credit Assignment in {RL} Training of {LLMs}},
  author={Kazemnejad, Amirhossein and Aghajohari, Milad and Portelance, Eva and Sordoni, Alessandro and Reddy, Siva and Courville, Aaron and Le Roux, Nicolas},
  booktitle={Proceedings of the 42nd International Conference on Machine Learning},
  series={PMLR},
  volume={267},
  pages={29557--29590},
  year={2025}
}

@article{qi2026bestpracticecriticoptimization,
  title={Best Practice Critic Optimization},
  author={Penghui Qi and Xiangxin Zhou and Wee Sun Lee},
  year={2026},
  journal={arXiv preprint arXiv:2608.23566}
}

@inproceedings{yu2025dapo,
  title={{DAPO}: An Open-Source {LLM} Reinforcement Learning System at Scale},
  author={Yu, Qiying and Zhang, Zheng and Zhu, Ruofei and others},
  booktitle={Advances in Neural Information Processing Systems},
  volume={38},
  year={2025}
}

@inproceedings{liu2025drgrpo,
  title={Understanding {R1-Zero}-Like Training: A Critical Perspective},
  author={Liu, Zichen and Chen, Changyu and Li, Wenjun and Qi, Penghui and Pang, Tianyu and Du, Chao and Lee, Wee Sun and Lin, Min},
  booktitle={Conference on Language Modeling},
  year={2025}
}

@inproceedings{ahmadian2024rloo,
  title={Back to Basics: Revisiting {REINFORCE}-Style Optimization for Learning from Human Feedback in {LLMs}},
  author={Ahmadian, Arash and Cremer, Chris and Gall{\'e}, Matthias and Fadaee, Marzieh and Kreutzer, Julia and Pietquin, Olivier and {\"U}st{\"u}n, Ahmet and Hooker, Sara},
  booktitle={Proceedings of the 62nd Annual Meeting of the Association for Computational Linguistics (Volume 1: Long Papers)},
  pages={12248--12267},
  year={2024},
  doi={10.18653/v1/2024.acl-long.662}
}

@inproceedings{li2026otb,
  title={The Optimal Token Baseline: Variance Reduction for Long-Horizon {LLM-RL}},
  author={Li, Yingru and Xu, Jiawei and Li, Ziniu and Liu, Jiacai and Liu, Wei and Tong, Yuxuan and Zheng, Longtao and Xue, Zhenghai and Zhang, Yaxiang and Cai, Tianle and Zhang, Ge and Liu, Qian and Wang, Baoxiang},
  year={2026},
  booktitle={Proceedings of the 43rd International Conference on Machine Learning},
  series={PMLR},
  volume={306},
  pages={71553--71575}
}

@article{deoliveira2025learningwithoutcritics,
  title={Learning Without Critics? {R}evisiting {GRPO} in Classical Reinforcement Learning Environments},
  author={de Oliveira, Bryan L. M. and Frujeri, Felipe V. and Queiroz, Marcos P. C. M. and Martins, Luana G. B. and Soares, Telma W. de L. and Melo, Luckeciano C.},
  journal={arXiv preprint arXiv:2511.03527},
  year={2025}
}

@article{sao2026,
  title={Single-Rollout Asynchronous Optimization for Agentic Reinforcement Learning},
  author={Zhenyu Hou and Yujiang Li and Jie Tang and Yuxiao Dong},
  year={2026},
  journal={arXiv preprint arXiv:2607.07508}
}

@article{evpo2026,
  title={{EVPO}: Explained Variance Policy Optimization for Adaptive Critic Utilization in {LLM} Post-Training},
  author={Chengjun Pan and Shichun Liu and Jiahang Lin and Dingwei Zhu and Jiazheng Zhang and Shihan Dou and Songyang Gao and Zhenhua Han and Binghai Wang and Rui Zheng and Xuanjing Huang and Tao Gui and Yansong Feng},
  journal={arXiv preprint arXiv:2604.19485},
  year={2026}
}

@article{zhang2026redippo,
  title={{ReDiPPO}: Reference-Guided Value Calibration and Discrepancy-Aware Token Reweighting for Mathematical Reasoning},
  author={Zhang, Zhenrong and Wu, Fei and Du, Jun and Zhang, Jianshu and Wei, Si},
  journal={arXiv preprint arXiv:2607.27631},
  year={2026}
}

@article{venkatraman2026lecritique,
  title={{Le Critique}: Privileged Value Functions for {LLM} Reinforcement Learning},
  author={Venkatraman, Siddarth and Dinot, Matthieu and Aitchison, Laurence},
  journal={arXiv preprint arXiv:2608.16739},
  year={2026}
}

@article{primeintellect2025intellect3,
  title={{INTELLECT-3}: Technical Report},
  author={{Prime Intellect Team}},
  journal={arXiv preprint arXiv:2512.16144},
  year={2025}
}

@misc{zhao2025icepop,
  title={Small Leak Can Sink a Great Ship---Boost {RL} Training on {MoE} with {IcePop}!},
  author={Zhao, Xin and Liu, Yongkang and Xu, Kuan and Guo, Jia and Wang, Zihao and Sun, Yan and Kong, Xinyu and Cao, Qianggang and Jiang, Liang and Wen, Zujie and Zhang, Zhiqiang and Zhou, Jun},
  year={2025},
  howpublished={\url{https://ringtech.notion.site/icepop}},
  note={Blog post}
}

@inproceedings{cheng2026isocompute,
  title={{IsoCompute} Playbook: Optimally Scaling Sampling Compute for {LLM} {RL}},
  author={Cheng, Zhoujun and Xie, Yutao and Qu, Yuxiao and Setlur, Amrith and Hao, Shibo and Pimpalkhute, Varad and Liang, Tongtong and Yao, Feng and Liu, Zhengzhong and Xing, Eric P. and Smith, Virginia and Salakhutdinov, Ruslan and Hu, Zhiting and Killian, Taylor W. and Kumar, Aviral},
  booktitle={Proceedings of the 43rd International Conference on Machine Learning},
  series={PMLR},
  volume={306},
  pages={19058--19078},
  year={2026}
}

@inproceedings{khatri2025artofscaling,
  title={The Art of Scaling Reinforcement Learning Compute for {LLMs}},
  author={Khatri, Devvrit and Madaan, Lovish and Tiwari, Rishabh and Bansal, Rachit and Duvvuri, Sai Surya and Zaheer, Manzil and Dhillon, Inderjit S. and Brandfonbrener, David and Agarwal, Rishabh},
  year={2026},
  booktitle={International Conference on Learning Representations}
}

@article{mccandlish2018largebatch,
  title={An Empirical Model of Large-Batch Training},
  author={McCandlish, Sam and Kaplan, Jared and Amodei, Dario and {OpenAI Dota Team}},
  journal={arXiv preprint arXiv:1812.06162},
  year={2018}
}

@inproceedings{ilyas2018closerlook,
  title={A Closer Look at Deep Policy Gradients},
  author={Ilyas, Andrew and Engstrom, Logan and Santurkar, Shibani and Tsipras, Dimitris and Janoos, Firdaus and Rudolph, Larry and Madry, Aleksander},
  booktitle={International Conference on Learning Representations},
  year={2020}
}

@article{li2026largerbatches,
  title={When Do Larger Batches Help Scale {LLM} Reinforcement Learning?},
  author={Li, Ziniu and Wang, Jinbo and Huang, Guanhua and Zhang, Feiyuan and Li, Pengbo and Chen, Alex},
  journal={arXiv preprint arXiv:2608.29296},
  year={2026}
}

@inproceedings{hilton2022batchinvariance,
  title={Batch Size-Invariance for Policy Optimization},
  author={Hilton, Jacob and Cobbe, Karl and Schulman, John},
  booktitle={Advances in Neural Information Processing Systems},
  year={2022},
  volume={35},
  pages={17086--17098}
}

@article{chen2025gpg,
  title={Group Policy Gradient},
  author={Chen, Junhua and Zhang, Zixi and Zhong, Hantao and Antonova, Rika},
  journal={arXiv preprint arXiv:2510.03679},
  year={2025}
}

@article{wang2025ragen,
  title={{RAGEN}: Understanding Self-Evolution in {LLM} Agents via Multi-Turn Reinforcement Learning},
  author={Wang, Zihan and Wang, Kangrui and Wang, Qineng and Zhang, Pingyue and Li, Linjie and Yang, Zhengyuan and Jin, Xing and Yu, Kefan and Nguyen, Minh Nhat and Liu, Licheng and Gottlieb, Eli and Lu, Yiping and Cho, Kyunghyun and Wu, Jiajun and Li, Fei-Fei and Wang, Lijuan and Choi, Yejin and Li, Manling},
  journal={arXiv preprint arXiv:2504.20073},
  year={2025}
}

@article{cui2025prime,
  title={Process Reinforcement through Implicit Rewards},
  author={Cui, Ganqu and Yuan, Lifan and Wang, Zefan and Wang, Hanbin and Zhang, Yuchen and Chen, Jiacheng and Li, Wendi and He, Bingxiang and Fan, Yuchen and Yu, Tianyu and Xu, Qixin and Chen, Weize and Yuan, Jiarui and Chen, Huayu and Zhang, Kaiyan and Lv, Xingtai and Wang, Shuo and Yao, Yuan and Han, Xu and Peng, Hao and Cheng, Yu and Liu, Zhiyuan and Sun, Maosong and Zhou, Bowen and Ding, Ning},
  journal={Transactions on Machine Learning Research},
  year={2026},
  url={https://openreview.net/forum?id=9SkkifLopZ}
}

@article{williams1992simple,
  title={Simple Statistical Gradient-Following Algorithms for Connectionist Reinforcement Learning},
  author={Williams, Ronald J.},
  journal={Machine Learning},
  volume={8},
  number={3--4},
  pages={229--256},
  year={1992},
  doi={10.1007/BF00992696},
  url={https://doi.org/10.1007/BF00992696}
}

@misc{pedregosa2026policygradient,
  title={Policy Gradients Part 1: The {REINFORCE} Estimator},
  author={Pedregosa, Fabian},
  year={2026},
  howpublished={\url{https://fa.bianp.net/blog/2026/policy-gradient/}},
  note={Blog post}
}

@article{greensmith2004variance,
  title={Variance Reduction Techniques for Gradient Estimates in Reinforcement Learning},
  author={Greensmith, Evan and Bartlett, Peter L. and Baxter, Jonathan},
  journal={Journal of Machine Learning Research},
  volume={5},
  pages={1471--1530},
  year={2004},
  url={https://jmlr.org/papers/v5/greensmith04a.html}
}

@inproceedings{li2025turnppo,
  title={{Turn-PPO}: Turn-Level Advantage Estimation with {PPO} for Improved Multi-Turn {RL} in Agentic {LLMs}},
  author={Li, Junbo and Zhou, Peng and Meng, Rui and Vadera, Meet P. and Li, Lihong and Li, Yang},
  booktitle={Findings of the Association for Computational Linguistics: EACL 2026},
  pages={6227--6243},
  year={2026},
  doi={10.18653/v1/2026.findings-eacl.328}
}

@inproceedings{kingma2015adam,
  title     = {Adam: A Method for Stochastic Optimization},
  author    = {Kingma, Diederik P. and Ba, Jimmy},
  booktitle = {International Conference on Learning Representations},
  year      = {2015}
}

@inproceedings{cutkosky2020momentum,
  title={Momentum Improves Normalized {SGD}},
  author={Cutkosky, Ashok and Mehta, Harsh},
  booktitle={Proceedings of the 37th International Conference on Machine Learning},
  pages={2260--2268},
  year={2020},
  series={PMLR},
  volume={119}
}

@inproceedings{loshchilov2019adamw,
  title     = {Decoupled Weight Decay Regularization},
  author    = {Loshchilov, Ilya and Hutter, Frank},
  booktitle = {International Conference on Learning Representations},
  year      = {2019}
}

@article{liu2025muon,
  title={{Muon} is Scalable for {LLM} Training},
  author={Liu, Jingyuan and Su, Jianlin and Yao, Xingcheng and others},
  journal={arXiv preprint arXiv:2502.16982},
  year={2025}
}

@inproceedings{merrill2026terminalbench,
  title={{Terminal-Bench}: Benchmarking Agents on Hard, Realistic Tasks in Command Line Interfaces},
  author={Merrill, Mike A. and Shaw, Alexander G. and Carlini, Nicholas and Li, Boxuan and Raj, Harsh and Bercovich, Ivan and Shi, Lin and Shin, Jeong Yeon and Walshe, Thomas and Buchanan, E. Kelly and others},
  year={2026},
  booktitle={International Conference on Learning Representations}
}

@misc{openthoughts2026tblite,
  title={{OpenThoughts-TBLite}: A High-Signal Benchmark for Iterating on Terminal Agents},
  author={{OpenThoughts-Agent team} and {Snorkel AI} and {Bespoke Labs}},
  year={2026},
  howpublished={\url{https://www.openthoughts.ai/blog/openthoughts-tblite}},
  note={Blog post}
}

@misc{deepscaler2025,
  title={{DeepScaleR}: Surpassing {O1-Preview} with a {1.5B} Model by Scaling {RL}},
  author={Michael Luo and Sijun Tan and Justin Wong and Xiaoxiang Shi and William Y. Tang and Manan Roongta and Colin Cai and Jeffrey Luo and Li Erran Li and Raluca Ada Popa and Ion Stoica},
  howpublished={\url{https://pretty-radio-b75.notion.site/DeepScaleR-Surpassing-O1-Preview-with-a-1-5B-Model-by-Scaling-RL-19681902c1468005bed8ca303013a4e2}},
  note={Blog post},
  year={2025}
}

@inproceedings{hendrycks2021math,
  title={Measuring Mathematical Problem Solving With the {MATH} Dataset},
  author={Hendrycks, Dan and Burns, Collin and Kadavath, Saurav and Arora, Akul and Basart, Steven and Tang, Eric and Song, Dawn and Steinhardt, Jacob},
  booktitle={Proceedings of the Neural Information Processing Systems Track on Datasets and Benchmarks},
  year={2021},
  volume={1}
}

@inproceedings{lightman2024verify,
  title     = {Let's Verify Step by Step},
  author    = {Lightman, Hunter and Kosaraju, Vineet and Burda, Yura and Edwards, Harri and Baker, Bowen and Lee, Teddy and Leike, Jan and Schulman, John and Sutskever, Ilya and Cobbe, Karl},
  booktitle = {International Conference on Learning Representations},
  year      = {2024}
}

@article{cobbe2021gsm8k,
  title   = {Training Verifiers to Solve Math Word Problems},
  author  = {Cobbe, Karl and Kosaraju, Vineet and Bavarian, Mohammad and Chen, Mark and Jun, Heewoo and Kaiser, Lukasz and Plappert, Matthias and Tworek, Jerry and Hilton, Jacob and Nakano, Reiichiro and Hesse, Christopher and Schulman, John},
  journal = {arXiv preprint arXiv:2110.14168},
  year    = {2021}
}

@inproceedings{lewkowycz2022minerva,
  title={Solving Quantitative Reasoning Problems with Language Models},
  author={Lewkowycz, Aitor and Andreassen, Anders and Dohan, David and Dyer, Ethan and Michalewski, Henryk and Ramasesh, Vinay and Slone, Ambrose and Anil, Cem and Schlag, Imanol and Gutman-Solo, Theo and Wu, Yuhuai and Neyshabur, Behnam and Gur-Ari, Guy and Misra, Vedant},
  booktitle={Advances in Neural Information Processing Systems},
  year={2022},
  volume={35},
  pages={3843--3857}
}

@inproceedings{he2024olympiadbench,
  title={{OlympiadBench}: A Challenging Benchmark for Promoting {AGI} with Olympiad-Level Bilingual Multimodal Scientific Problems},
  author={He, Chaoqun and Luo, Renjie and Bai, Yuzhuo and Hu, Shengding and Thai, Zhen Leng and Shen, Junhao and Hu, Jinyi and Han, Xu and Huang, Yujie and Zhang, Yuxiang and Liu, Jie and Qi, Lei and Liu, Zhiyuan and Sun, Maosong},
  booktitle={Proceedings of the 62nd Annual Meeting of the Association for Computational Linguistics (Volume 1: Long Papers)},
  year={2024},
  pages={3828--3850},
  doi={10.18653/v1/2024.acl-long.211}
}

@misc{cao2025skyrl,
  title  = {{SkyRL-v0}: Train Real-World Long-Horizon Agents via Reinforcement Learning},
  author = {Shiyi Cao and Sumanth Hegde and Dacheng Li and Tyler Griggs and Shu Liu and Eric Tang and Jiayi Pan and Xingyao Wang and Akshay Malik and Graham Neubig and Kourosh Hakhamaneshi and Richard Liaw and Philipp Moritz and Matei Zaharia and Joseph E. Gonzalez and Ion Stoica},
  year   = {2025},
  howpublished = {\url{https://github.com/NovaSky-AI/SkyRL}}
}

@misc{griggs2025skyrl,
  title  = {Evolving {SkyRL} into a Highly-Modular {RL} Framework},
  author = {Tyler Griggs and Sumanth Hegde and Eric Tang and Shu Liu and Shiyi Cao and Dacheng Li and Charlie Ruan and Philipp Moritz and Kourosh Hakhamaneshi and Richard Liaw and Akshay Malik and Matei Zaharia and Joseph E. Gonzalez and Ion Stoica},
  year   = {2025},
  howpublished = {\url{https://novasky-ai.notion.site/skyrl-v01}},
  note   = {Blog post}
}

@inproceedings{zhao2011analysis,
  title={Analysis and Improvement of Policy Gradient Estimation},
  author={Zhao, Tingting and Hachiya, Hirotaka and Niu, Gang and Sugiyama, Masashi},
  booktitle={Advances in Neural Information Processing Systems}, volume={24}, year={2011}}

@article{papini2022smoothing,
  title={Smoothing Policies and Safe Policy Gradients},
  author={Papini, Matteo and Pirotta, Matteo and Restelli, Marcello},
  journal={Machine Learning}, volume={111}, number={11}, pages={4081--4137}, year={2022}, doi={10.1007/s10994-022-06232-6}}

@inproceedings{yin2018gradientdiversity,
  title={Gradient Diversity: a Key Ingredient for Scalable Distributed Learning},
  author={Yin, Dong and Pananjady, Ashwin and Lam, Max and Papailiopoulos, Dimitris and Ramchandran, Kannan and Bartlett, Peter L.},
  booktitle={Proceedings of the 21st International Conference on Artificial Intelligence and Statistics},
  series={PMLR},
  volume={84},
  pages={1998--2007},
  year={2018}
}

@inproceedings{wu2018variance,
  title={Variance Reduction for Policy Gradient with Action-Dependent Factorized Baselines},
  author={Wu, Cathy and Rajeswaran, Aravind and Duan, Yan and Kumar, Vikash and Bayen, Alexandre M. and Kakade, Sham M. and Mordatch, Igor and Abbeel, Pieter},
  booktitle={International Conference on Learning Representations},
  year={2018}
}

@article{qwen2025qwen25,
  title={{Qwen2.5} Technical Report},
  author={{Qwen} and Yang, An and Yang, Baosong and Zhang, Beichen and Hui, Binyuan and Zheng, Bo and Yu, Bowen and others},
  journal={arXiv preprint arXiv:2412.15115},
  year={2025}
}

@misc{qwen2026qwen35,
  title={{Qwen3.5}: Towards Native Multimodal Agents},
  author={{Qwen Team}},
  year={2026},
  howpublished={\url{https://qwen.ai/blog?id=qwen3.5}},
  note={Blog post}
}

@inproceedings{kwon2023vllm,
  title={Efficient Memory Management for Large Language Model Serving with {PagedAttention}},
  author={Kwon, Woosuk and Li, Zhuohan and Zhuang, Siyuan and Sheng, Ying and Zheng, Lianmin and Yu, Cody Hao and Gonzalez, Joseph E. and Zhang, Hao and Stoica, Ion},
  booktitle={Proceedings of the 29th Symposium on Operating Systems Principles},
  year={2023},
  pages={611--626},
  doi={10.1145/3600006.3613165}
}

@inproceedings{todorov2012mujoco,
  title={{MuJoCo}: A physics engine for model-based control},
  author={Todorov, Emanuel and Erez, Tom and Tassa, Yuval},
  booktitle={IEEE/RSJ International Conference on Intelligent Robots and Systems},
  pages={5026--5033},
  year={2012}
}

@misc{anonymous2026measuring,
  title        = {Measuring Gradient Geometry in Language-Model {RLVR} with Sketched Gradients},
  author       = {Anonymous},
  year         = {2026},
  howpublished = {\url{https://openreview.net/forum?id=7tjAyMuPu9}},
  note         = {Under review at the Fifteenth International Conference on Learning Representations}
}

@inproceedings{kakade2001natural,
  title={A Natural Policy Gradient},
  author={Kakade, Sham M.},
  booktitle={Advances in Neural Information Processing Systems},
  volume={14},
  year={2001}
}

@article{williams1991function,
  title={Function Optimization Using Connectionist Reinforcement Learning Algorithms},
  author={Williams, Ronald J. and Peng, Jing},
  journal={Connection Science},
  volume={3},
  number={3},
  pages={241--268},
  year={1991}
}

@inproceedings{wang2026sppo,
  title={{SPPO}: Sequence-Level {PPO} for Long-Horizon Reasoning Tasks},
  author={Wang, Tianyi and Li, Yixia and Li, Long and Chen, Yibiao and Huang, Shaohan and Chen, Yun and Li, Peng and Liu, Yang and Chen, Guanhua},
  booktitle={Proceedings of the 64th Annual Meeting of the Association for Computational Linguistics (Volume 1: Long Papers)},
  pages={7831--7853},
  year={2026},
  doi={10.18653/v1/2026.acl-long.356}
}

@article{liang2026sapo,
  title={{SAPO}: Single-Rollout Autoregressive Policy Optimization for Agentic Reinforcement Learning},
  author={Liang, Dayang and Feng, Lang and An, Bo and Liu, Yunlong},
  journal={arXiv preprint arXiv:2608.19842},
  year={2026}
}

@inproceedings{liu2026srpo,
  title={{SRPO}: Self-Reflective Policy Optimization for Long-Horizon Reasoning},
  author={Liu, Jialong and Shi, Yuling and Yang, Ning and Gu, Xiaodong and Li, Zuchao},
  booktitle={Proceedings of the 43rd International Conference on Machine Learning},
  year={2026},
  series={PMLR},
  volume={306},
  pages={78359--78388}
}

@article{chen2026respo,
  title={{ReSPO}: Reshaped Sequence Policy Optimization for Gradient Starvation in Off-Policy Learning},
  author={Chen, Yihang and Ban, Yuanhao and Hsieh, Cho-Jui},
  journal={arXiv preprint arXiv:2609.35433},
  year={2026}
}

@inproceedings{charikar2002countsketch,
  title={Finding Frequent Items in Data Streams},
  author={Charikar, Moses and Chen, Kevin and Farach-Colton, Martin},
  booktitle={Proceedings of the 29th International Colloquium on Automata, Languages and Programming},
  pages={693--703},
  year={2002}
}

\clearpage
\appendix

\section{Theory}
\label{app:theory}

\subsection{Length-Adaptive GAE}
\label{app:gae-derivation}

Ignoring the endpoint, the weights of Equation~\ref{eq:gae} form a geometric
distribution over $k$ with mean $1/(1-\lambda)$. Under length-adaptive GAE
(Equation~\ref{eq:length-adaptive-gae}), this decay scale is $\alpha H$, and a
signal $d=fH$ transitions ahead keeps relative weight
\begin{equation}
\lambda^{d}=\left(1-\frac{1}{\alpha H}\right)^{fH}\approx e^{-f/\alpha}.
\end{equation}
Thus $\alpha$ sets the decay scale relative to the trajectory length. At the
first token of a long response, $f\approx1$, so with terminal-only rewards and
$\gamma=1$, the Monte Carlo return keeps weight about $e^{-1/\alpha}$, the
values given in Section~\ref{sec:ppo-background}.  The expression for $\lambda(H)$ lies in
$(0,1)$ only when $\alpha H>1$. Shorter trajectories require clamping
$\lambda$, which we set to zero in the control tasks
(Appendix~\ref{app:control-networks}).

\subsection{Gradient Variance Scaling}
\label{app:gradient-scaling}

\begin{proposition}[Monte Carlo gradient variance]
\label{prop:mc-gradient-variance}
Consider trajectories from a fixed policy and task distribution.
Suppose $|r(\tau)|\le M_H$ and
$\mathbb E[\|\nabla_\theta\log\pi_\theta(a_t\mid s_t)\|^2]\le C$
for every $t$, where $C$ is independent of $H$. Then
\begin{equation}
V=\operatorname{Var}(\hat g)\le CM_H^2H.
\label{eq:mc-gradient-variance}
\end{equation}
In particular, accumulated rewards with $|r_t|\le r_{\max}$ give
$M_H=Hr_{\max}$ and $V\le Cr_{\max}^2H^3=O(H^3)$.
\end{proposition}

The variance bound of Proposition~\ref{cor:outcome-gradient-variance} is the
bounded-outcome case $M_H=1$.

\paragraph{Proof of Propositions~\ref{prop:mc-gradient-variance}
and~\ref{cor:outcome-gradient-variance}.}
The argument follows the score-function variance calculation of
\citet[Appendix~D, Lemma~28]{papini2022smoothing} and the exposition of
\citet{pedregosa2026policygradient}. Assume a differentiable policy with
parameter-independent support and interchangeable differentiation and
integration, with rewards, start states, and transitions independent of
$\theta$, and write $\operatorname{Var}(Y)=\mathbb E\|Y-\mathbb E Y\|^2$ and
$X_t=\nabla_\theta\log\pi_\theta(a_t\mid s_t)$. The trajectory density
satisfies $\nabla_\theta p_\theta(\tau)=p_\theta(\tau)\sum_tX_t$, so
differentiating $\mathbb E[r(\tau)]$ gives $\mathbb E[\hat g]=g$. Let
$h_t=(s_1,a_1,\ldots,s_t)$ be the history before $a_t$. Because the policy
depends only on $s_t$ and action probabilities integrate to one,
$\mathbb E[X_t\mid h_t]=\nabla_\theta\int\pi_\theta(a\mid s_t)\,\mathrm da=0$.
For $j<t$, $X_j$ is determined by $h_t$, so
$\mathbb E[X_j^\top X_t]=\mathbb E[X_j^\top\mathbb E[X_t\mid h_t]]=0$, and the
cross terms vanish:
\begin{equation}
\mathbb E\!\left[\left\|\sum_{t=1}^{H}X_t\right\|^2\right]
=\sum_{t=1}^{H}\mathbb E\|X_t\|^2\le CH.
\end{equation}
Bounding the squared return before taking the expectation gives
\begin{equation}
\operatorname{Var}(\hat g)
\le\mathbb E\!\left[r(\tau)^2\left\|\sum_{t=1}^{H}X_t\right\|^2\right]
\le CM_H^2H,
\end{equation}
and independent trajectories divide this bound by $B$. Taking
$M_H=Hr_{\max}$ gives $V\le Cr_{\max}^2H^3$, where one factor of $H$ comes from
the summed action gradients and two from squaring the accumulated return. This
recovers the known $O(H^3)$
bound~\citep{zhao2011analysis,papini2022smoothing}. Taking $M_H=1$
proves the variance bound of
Proposition~\ref{cor:outcome-gradient-variance}. For the noise scale, the same
calculation bounds the second moment, $\mathbb E\|\hat g\|^2\le CH$, so
\begin{equation}
\nu=\frac{\mathbb E\|\hat g\|^2-\|g\|^2}{\|g\|^2}\le\frac{CH}{\|g\|^2}-1,
\end{equation}
and $\|g\|\ge m$ gives $\nu\le CH/m^2-1=O(H)$.

\paragraph{Relation to our objectives.}
The calculation uses summed Monte Carlo gradients. For fixed-length
trajectories, averaging over policy tokens divides the estimate by $H$, which
rescales its mean and standard deviation alike and leaves its relative error
unchanged when the mean is nonzero. The bound also applies to
rewards minus a fixed per-prompt baseline when both lie in $[0,1]$,
since such a baseline preserves the expected
gradient~\citep{williams1992simple,greensmith2004variance}. It does not cover
variable-length token averaging, GRPO's group baseline, PPO's bootstrapped
advantages, or the masking of our shared objective. It motivates varying the batch size with the horizon, but does not
by itself predict where PPO and GRPO cross.

\subsection{The Cosine Model and the Crossover Batch Size}
\label{app:alignment}

\paragraph{Setup.}
Fix the policy, and let $J(\theta)=\mathbb E_{\tau\sim\pi_\theta}[r(\tau)]$, so
that $g=\nabla_\theta J(\theta)$. The batch gradient $\hat g_B$ averages $B$
independent per-trajectory estimates with mean $\mu=\mathbb E[\hat g]$ and
covariance $\Sigma$. Write $\hat g_B=\mu+\varepsilon$, where
$\mathbb E[\varepsilon]=0$ and $\operatorname{Cov}(\varepsilon)=\Sigma/B$. As in
the main text, $c=\cos(\mu,g)$ and $\nu=\operatorname{tr}\Sigma/\|\mu\|^2$. We
assume $g$ and $\mu$ are nonzero and that $\hat g$ has finite fourth moments,
which the error bound in Equation~\ref{eq:cosine-error} requires. The moment
condition holds in our settings, since rewards, advantages, and response lengths are bounded in
the LLM tasks and the Gaussian policy's score has all moments in the control
tasks. The cross term has zero mean, so
\begin{equation}
\mathbb E\|\hat g_B\|^2
=\|\mu\|^2+\mathbb E\|\varepsilon\|^2
=\|\mu\|^2\left(1+\nu/B\right).
\label{eq:second-moment}
\end{equation}

\paragraph{Progress per update.}
Suppose $J$ is $\beta$-smooth, so that
$|J(\theta+u)-J(\theta)-\langle g,u\rangle|\le\tfrac{\beta}{2}\|u\|^2$. A step
of fixed length, $u=\eta\,\hat g_B/\|\hat g_B\|$, has
$\langle g,u\rangle=\eta\|g\|\cos(\hat g_B,g)$ and $\|u\|=\eta$, so
\begin{equation}
\bigl|\mathbb E[\Delta J]-\eta\|g\|\,\mathbb E[\cos(\hat g_B,g)]\bigr|
\le\tfrac{\beta}{2}\eta^2.
\end{equation}
At a fixed policy, the factor $\eta\|g\|$ is shared by all estimators, so to
first order in $\eta$ they are ranked by their expected cosine. The ranking is
guaranteed when the expected cosines differ by more than $\beta\eta/\|g\|$.
The ranking concerns steps along the batch gradient itself. It does not cover
update directions that intentionally depart from $g$. Natural policy gradients
and adaptive optimizers precondition the
gradient~\citep{kakade2001natural,kingma2015adam}, and entropy bonuses and asymmetric
clipping change the objective to encourage
exploration~\citep{williams1991function,yu2025dapo}. For such methods, the
expected cosine measures how well an estimator recovers $g$, not how much the
resulting update improves $J$.

\paragraph{Expected cosine.}
Let $N=\langle\hat g_B,g\rangle/\|g\|$, $D=\|\hat g_B\|$, and
$\bar D=(\mathbb E D^2)^{1/2}$, and set $\cos(0,g)=0$, since GRPO gives
$\hat g_B=0$ when every group has identical rewards. The numerator is linear,
so $\mathbb E N=c\|\mu\|$, and Equation~\ref{eq:second-moment} gives
$\mathbb E N/\bar D=c/\sqrt{1+\nu/B}$. Because $N/D=\cos(\hat g_B,g)$, the gap
between the cosine model and the expected cosine is
$\mathbb E[\cos(\hat g_B,g)(D/\bar D-1)]$. Using $|\cos|\le1$ and
$|x-1|\le|x^2-1|$ for $x\ge0$,
\begin{equation}
\left|\mathbb E\!\left[\cos(\hat g_B,g)\right]-\frac{c}{\sqrt{1+\nu/B}}\right|
\le\frac{\sqrt{\operatorname{Var}\|\hat g_B\|^2}}{\mathbb E\|\hat g_B\|^2}.
\label{eq:cosine-error}
\end{equation}
The cosine model is therefore accurate whenever the squared norm of the batch
gradient concentrates, and Appendix~\ref{app:control-checks} compares it with
measured cosines. As $B\to\infty$, $\hat g_B\to\mu$ almost surely, and since
the cosine is bounded and continuous at $\mu\neq0$,
$\mathbb E[\cos(\hat g_B,g)]\to c$ by dominated convergence. Like the
squared-bias floor, this limit does not depend on $\nu$, so more samples cannot
remove it.

\paragraph{Group-mean baseline.}
With a group-mean baseline that includes the response itself, response $i$
receives advantage $(1-1/G)r_i-(1/G)\sum_{j\neq i}r_j$. Given the prompt, the
other responses are independent of response $i$, and its score function has zero
conditional mean, so the second term contributes nothing in expectation. Hence
$\mathbb E[\hat g]=(1-1/G)\,g$, which has $c=1$ but squared bias
$\|g\|^2/G^2$. Squared error therefore counts this estimator as biased even
though its mean points exactly along $g$~\citep{ahmadian2024rloo,liu2025drgrpo}.

\paragraph{Crossover batch size.}
Consider two estimators with alignments $c_1>c_2>0$ and noise scales
$\nu_1,\nu_2$. The main text takes GRPO as estimator 1 and PPO as
estimator 2. Squaring preserves the order of positive cosines, so under the
cosine model, estimator 1 has the larger expected cosine when
$c_1^2/(1+\nu_1/B)>c_2^2/(1+\nu_2/B)$. Multiplying by
$(B+\nu_1)(B+\nu_2)/B>0$ shows this is equivalent to
\begin{equation}
B>B^*_{\cos}=\frac{c_2^2\nu_1-c_1^2\nu_2}{c_1^2-c_2^2}.
\end{equation}
The condition is linear in $B$, so the two curves cross at most once.
Estimator 2 leads for every $B<B^*_{\cos}$ and estimator 1 for every
$B>B^*_{\cos}$. If $B^*_{\cos}\le0$, estimator 1 leads at every batch size.
This proves Equation~\ref{eq:crossover-batch}. The same form applies when
$c_1<1$, as for GRPO with standard-deviation normalization in our LLM
settings. Scaling $\nu_1$ and $\nu_2$ by a common factor scales $B^*_{\cos}$ by
the same factor. If both noise scales grew in proportion to $H$, which
Proposition~\ref{cor:outcome-gradient-variance} permits but does not imply,
the crossover would move to proportionally larger batch sizes.

\paragraph{Parity.}
When the alignments are close, $c_{\mathrm{PPO}}\approx c_{\mathrm{GRPO}}$, the
denominator of Equation~\ref{eq:crossover-batch} is small and $B^*_{\cos}$ is
large. At large batch sizes, both expected cosines then approach nearly the
same ceiling, so neither estimator has a clear advantage and training can end
tied. If the alignments are equal and PPO has the lower noise scale, the two
curves approach the same limit without crossing.

\paragraph{Dependent trajectories.}
GRPO's trajectories share a group baseline and are not independent within a
group. The results above hold with the $P=B/G$ independent groups as units. If
$\Sigma_G$ is the covariance of one group's average gradient, then
$\operatorname{Cov}(\hat g_B)=G\Sigma_G/B$, so defining
$\nu_{\mathrm{GRPO}}=G\operatorname{tr}\Sigma_G/\|\mu\|^2$ leaves $\nu/B$
unchanged. This $\nu$ depends on $G$. PPO's advantage whitening over the batch
also couples trajectories slightly. The cosine model treats them as
independent, and Appendix~\ref{app:control-checks} shows that it still matches
the measured cosines of both estimators.

\clearpage
\section{Experimental Details}
\label{app:details}

\subsection{Control Tasks}
\label{app:control-networks}
\label{app:pendulum}
\label{app:halfcheetah}

Table~\ref{tab:control-settings} lists the control settings.

\begin{table*}[t]
\centering
\small
\renewcommand{\arraystretch}{1.15}
\begin{tabularx}{\textwidth}{@{}p{3.1cm}>{\raggedright\arraybackslash}X@{}}
\toprule
Setting & Value \\
\midrule
Environments & Gymnasium Pendulum-v1 and HalfCheetah-v4 with default start states, fixed horizon $H\in\{5,25,200\}$, no early termination \\
Reward & accumulated per-step reward, rescaled to the score $S(\tau)$ and given only at the terminal state \\
Batch sizes $B$ & powers of four, $4$--$4{,}096$ at $H=5$, $16$--$16{,}384$ at $H=25$, and $16$--$65{,}536$ at $H=200$ \\
Group size $G$ & GRPO $16$ ($4$ at $B=4$), PPO $1$ \\
Training & 5,000 updates, five seeds \\
Actor & two-layer MLP, width 64, $\tanh$ activations, Gaussian mean with one state-independent log standard deviation per action dimension, $\tanh$-squashed and scaled to the action range \\
Critic (PPO) & separate network with the same architecture and a scalar output \\
Initialization & orthogonal, gain $\sqrt2$ for hidden layers and $0.01$ for the action-mean head, log standard deviation $0$, critic output weights $0$ and bias $0.63$ (Pendulum) or $0.1214$ (HalfCheetah), identical actors for PPO and GRPO at each seed \\
Actor update & one unit-norm SGD step per batch, $\eta=10^{-3}$, no momentum, weight decay, or annealing \\
Critic update & one full-batch unit-norm step of size $5\times10^{-3}$ after each actor step, on the squared error to $S(\tau)$ at every transition ($\lambda_V=1$), no value clipping or pretraining \\
PPO advantages & GAE with $\gamma=1$ and length-adaptive $\lambda$ clamped at $0$, $\alpha\in\{0.1,0.5,1.5\}$, normalized once over the batch's transitions with the population standard deviation plus $10^{-8}$ \\
GRPO advantages & return minus the inclusive group mean (Dr.~GRPO). GRPO (std) also divides by the group's population standard deviation plus $10^{-8}$ and zeroes groups whose standard deviation is at most $10^{-8}$ \\
Additional variants & PPO with $\lambda=1$ and GRPO (std) (Appendix~\ref{app:all-control}) \\
Evaluation & every 10 updates, four stochastic trajectories from each of 1,024 validation states shared across methods, best score per run \\
\bottomrule
\end{tabularx}
\caption{\textbf{Control-task settings.}}
\label{tab:control-settings}
\end{table*}

\paragraph{Pendulum.}
The policy applies torque to swing up and stabilize a pole.  The
actor receives $x_t=[\cos\theta_t,\ \sin\theta_t,\ \dot\theta_t/8,\ t/H]$ and
outputs one torque $u_t\in[-2,2]$. With the standard per-step reward
$r_t=-(\operatorname{angle}(\theta_t)^2+0.1\dot\theta_t^2+0.001u_t^2)$ and
return $R(\tau)=\sum_tr_t$, the score
\begin{equation}
S(\tau)=1+\frac{R(\tau)}{H(\pi^2+0.1\cdot8^2+0.001\cdot2^2)}
\end{equation}
lies in $[0,1]$ by the reward bounds.

\paragraph{HalfCheetah.}
The policy controls a planar articulated runner, simulated with the Gymnasium
MuJoCo model~\citep{todorov2012mujoco}.  The actor observes the
eight positions other than absolute horizontal position $p_t$, all nine velocities,
and $t/H$, with physical features divided by 10 and clipped to $[-1,1]$, and it
outputs six controls in $[-1,1]$. HalfCheetah therefore has 18 input and 6
action dimensions, against 4 and 1 for Pendulum. The standard per-step reward
$r_t=(p_{t+1}-p_t)/0.05-0.1\sum_{j=1}^{6}a_{t,j}^2$ rewards forward movement and
penalizes control effort. With a per-step reference range of $[-2,12]$, the
score is $S(\tau)=(R(\tau)+2H)/(14H)$, and we never observe a score outside
$[0,1]$.

\paragraph{Critic inputs and updates.}
PPO's critic receives the actor's inputs together with the reward accumulated
strictly before the current action, under the same affine scaling as the score.
This feature summarizes the relevant trajectory history, allowing a
feed-forward critic to predict the final score without recurrence. It excludes
the current and future rewards, so it is valid privileged information
(Section~\ref{sec:ppo-background}), and the actor does not use it. PPO computes
GAE with the critic before fitting the critic to the current batch.

\paragraph{Evaluation.}
Taking each run's best of 500 evaluations raises scores by about
$5\times10^{-4}$ over the mean of its last ten, by similar amounts for PPO and
GRPO. Because PPO and GRPO share the initial actor and validation states at
each seed, we compute the standard errors of
Section~\ref{sec:comparison-protocol} from per-seed differences.

\subsection{LLM Tasks}
\label{app:llm-data}

Table~\ref{tab:llm-settings} lists the LLM settings. We swept the critic
learning rate over $\{2.5\times10^{-6},5\times10^{-6}\}$ and the number of
critic epochs per batch over $\{1,2\}$, and used one epoch with the learning
rates in the table. The actor learning rate of $10^{-6}$ follows prior RL
fine-tuning of LLMs~\citep{deepscaler2025,yu2025dapo}. GRPO and PPO come from
independent runs, so the standard error of their difference is
$\mathrm{SE}_{\mathrm{diff}}=\sqrt{\mathrm{SE}_{\mathrm{GRPO}}^2+\mathrm{SE}_{\mathrm{PPO}}^2}$.
The LLM experiments use SkyRL~\citep{cao2025skyrl,griggs2025skyrl} and vLLM~\citep{kwon2023vllm}.

\begin{table*}[t]
\centering
\small
\renewcommand{\arraystretch}{1.15}
\setlength{\tabcolsep}{7pt}
\begin{tabularx}{\textwidth}{@{}p{3.1cm}>{\raggedright\arraybackslash}X>{\raggedright\arraybackslash}X@{}}
\toprule
Setting & Math & Terminal agents \\
\midrule
Policy & Qwen2.5-3B (base) & Qwen3.5-4B \\
Data & 10,000 DeepScaleR training and 500 validation questions & 14,101 TMax-15K training and 500 validation tasks \\
Batch sizes $B$ & $4$, $16$, $64$, $256$, $1{,}024$, $4{,}096$ & $128$, $512$, $2{,}048$ \\
Group size $G$ & GRPO $\min(16,\sqrt B)$, PPO $1$ & GRPO $16$, PPO $1$ \\
Steps and seeds & 400 steps, three seeds & 200 steps, one seed \\
Updates per step & two actor updates on $2B$ trajectories, the second one step off-policy & one actor update, asynchronous, at most four policy versions stale \\
Length limits & 1,024 prompt and 4,096 response tokens & 64 turns, 65,536-token context, 32,768 tokens per turn \\
Critic learning rate & $2.5\times10^{-6}$, one update per actor update & $5\times10^{-6}$, one update per step \\
Critic pretraining & 200 updates of 16 base-policy trajectories, learning rate $2\times10^{-6}$ & 250 updates of 64 initial-policy trajectories, learning rate $2\times10^{-6}$ \\
Privileged input & reference answer and solution & grading tests and target state \\
Hardware per run & 2--8 H100 GPUs & 8--24 H100 GPUs, trainer and inference \\
\midrule
\multicolumn{3}{@{}l}{\textit{Both settings}} \\
\midrule
Optimizer & \multicolumn{2}{>{\raggedright\arraybackslash\hsize=\dimexpr2\hsize+2\tabcolsep\relax}X@{}}{AdamW, $\beta=(0.9,0.999)$, weight decay $0.01$, gradient clipping at norm $1.0$} \\
Actor learning rate & \multicolumn{2}{>{\raggedright\arraybackslash\hsize=\dimexpr2\hsize+2\tabcolsep\relax}X@{}}{$10^{-6}$, 20-step linear warmup, no decay, no Kullback--Leibler or entropy term} \\
Sampling & \multicolumn{2}{>{\raggedright\arraybackslash\hsize=\dimexpr2\hsize+2\tabcolsep\relax}X@{}}{temperature $1.0$, top-$p$ $1.0$} \\
GRPO advantages & \multicolumn{2}{>{\raggedright\arraybackslash\hsize=\dimexpr2\hsize+2\tabcolsep\relax}X@{}}{group mean and standard-deviation normalization} \\
PPO advantages & \multicolumn{2}{>{\raggedright\arraybackslash\hsize=\dimexpr2\hsize+2\tabcolsep\relax}X@{}}{GAE with $\gamma=1$, length-adaptive $\lambda$, and $\lambda_V=1$, whitened over the batch} \\
Critic & \multicolumn{2}{>{\raggedright\arraybackslash\hsize=\dimexpr2\hsize+2\tabcolsep\relax}X@{}}{policy backbone with a scalar value head, token-mean squared error, no value clipping, full-attention sublayers frozen during RL} \\
\bottomrule
\end{tabularx}
\caption{\textbf{LLM settings.} PPO's critic is initialized from the
pretrained critic, which all PPO runs in a setting share.}
\label{tab:llm-settings}
\end{table*}

\paragraph{Math data and evaluation.}
We estimate question difficulty by sampling Qwen2.5-3B Instruct five times per
DeepScaleR question and keep questions with one to four correct answers,
taking 2,500 training and 125 validation questions from each of these four
buckets. The prompt asks the model to reason step by step and put the final
answer in \verb|\boxed{}|. The reward is $1$ for a correct boxed answer and $0$
otherwise. During training, we score the validation set with four samples per
question every 20 steps. After training, we re-score every checkpoint within
2 points of a run's best with 16 samples per question and select the best. At
the selected checkpoint, we evaluate MATH-500 (8 samples per question), GSM8K
(4), AMC23 (128), Minerva Math (16), and OlympiadBench (8), at temperature
$1.0$ with a 4,096-token limit, and report mean sampled accuracy. These
benchmarks have 500, 1,319, 40, 272, and 581 problems, the last being the
problems of OlympiadBench's English text-only math set that have a single
final answer.

\paragraph{Terminal environment and evaluation.}
The agent issues one \texttt{bash} tool call per turn in a persistent shell,
with a 120-second command timeout and tool outputs truncated to 10,000
characters. An episode ends when the agent submits, or with reward $0$ at a
format error or when a turn, context, or per-turn token limit is reached. On
submission, the task's hidden test script runs in the sandbox, and its score,
clipped to $[0,1]$, is the reward. Model-generated commands run in isolated
Modal sandboxes. Settings not listed in Table~\ref{tab:llm-settings} follow
Tmax~\citep{ivison2026tmax}, except that we mask groups with identical rewards
out of the loss instead of resampling them and run the language-model head in
bf16. Terminal-Bench
2.1 has 89 tasks, and we use 98 of TB-Lite's 100 tasks. From TB-Lite, we exclude \texttt{git-repo-forensics}, whose required repository is missing, and \texttt{neutron-submission}, whose missing input data allow an empty report to receive full credit. Test scores are mean
reward with one standard error over eight evaluation passes, each with one
attempt per task, and the benchmark average is the mean of the two benchmark
scores in each pass.

\paragraph{Compute.}
The GPU-hours in Figure~\ref{fig:short-math-compute} estimate the GPU time an
efficient implementation of each run needs over its step budget. Both settings
count four components. Generation is the GPU time to produce the trained
trajectories. Actor training counts forward and backward passes and optimizer
steps over the tokens that receive a gradient, so GRPO skips groups whose
rewards are identical. Critic training counts the same for PPO's critic, whose
forward pass also supplies the values for GAE. Weight synchronization counts
copying the updated actor to the inference engines. The masked objective uses
the inference engine's log-probabilities, so neither method needs a separate
log-probability pass. In math, generation and training alternate on the same
GPUs, so we measure generation time directly. In terminal agents, training is
asynchronous, and depending on how GPUs are divided between the trainer and the
inference engines, either side can sit idle while waiting for the other. We
therefore charge every run's generated tokens at 1,583 tokens per second per
H100, the highest throughput in an offline replay of 192 of our trajectories
under the training limits, reached with 48 trajectories per engine as in
training. The replay includes the measured 2.6 seconds of sandbox time per
turn.  Both settings exclude evaluation, checkpointing, idle time, and the
one-time critic pretraining that all PPO runs in a setting share, which costs
about $0.5$ GPU-hours in math and about $50$ in terminal agents. In total, we
estimate that the LLM experiments used about 21,000 H100 GPU-hours, including
critic pretraining, evaluation, tuning, and diagnostics.

\clearpage
\section{Additional Results}
\label{app:additional-results}

\subsection{Full LLM Results}

Tables~\ref{tab:math-all-scores} and~\ref{tab:terminal-all-scores} list every
math and terminal-agent score, including the validation scores.

\begin{table*}[t]
\centering

\footnotesize
\setlength{\tabcolsep}{4pt}
\begin{tabular}{@{}llccccccc@{}}
\toprule
$B$ & Method & DeepScaleR & MATH-500 & GSM8K & AMC23 & Minerva & OlympiadBench & Avg. \\
\midrule
4 & GRPO & 35.0\,$\pm$\,1.0 & 57.1\,$\pm$\,0.8 & 77.2\,$\pm$\,0.6 & 30.4\,$\pm$\,0.7 & 13.5\,$\pm$\,0.7 & 23.3\,$\pm$\,0.4 & 40.3\,$\pm$\,0.5 \\
 & PPO $\alpha=0.1$ & \textbf{39.5\,$\pm$\,2.1} & 58.8\,$\pm$\,1.5 & 80.0\,$\pm$\,1.7 & 32.3\,$\pm$\,1.9 & \textbf{15.9\,$\pm$\,0.7} & 25.0\,$\pm$\,0.7 & 42.4\,$\pm$\,1.2 \\
 & PPO $\alpha=0.5$ & 39.3\,$\pm$\,0.6 & \textbf{58.9\,$\pm$\,0.7} & \textbf{81.1\,$\pm$\,0.8} & \textbf{33.9\,$\pm$\,1.2} & 15.5\,$\pm$\,0.5 & \textbf{25.6\,$\pm$\,0.2} & \textbf{43.0\,$\pm$\,0.6} \\
 & PPO $\alpha=1.5$ & 37.6\,$\pm$\,0.7 & 58.1\,$\pm$\,0.8 & 78.6\,$\pm$\,0.3 & 31.4\,$\pm$\,1.2 & 14.5\,$\pm$\,0.2 & 25.0\,$\pm$\,0.8 & 41.5\,$\pm$\,0.6 \\
\addlinespace
16 & GRPO & 41.2\,$\pm$\,0.4 & 59.8\,$\pm$\,0.2 & 81.0\,$\pm$\,0.4 & 34.2\,$\pm$\,0.8 & 15.7\,$\pm$\,0.6 & 26.0\,$\pm$\,0.1 & 43.3\,$\pm$\,0.3 \\
 & PPO $\alpha=0.1$ & \textbf{45.5\,$\pm$\,1.1} & 60.8\,$\pm$\,0.6 & \textbf{82.5\,$\pm$\,0.3} & 35.5\,$\pm$\,0.7 & \textbf{17.0\,$\pm$\,0.4} & \textbf{27.5\,$\pm$\,0.5} & 44.7\,$\pm$\,0.3 \\
 & PPO $\alpha=0.5$ & 44.3\,$\pm$\,0.9 & \textbf{62.2\,$\pm$\,0.2} & 82.3\,$\pm$\,1.0 & \textbf{37.7\,$\pm$\,0.2} & 16.4\,$\pm$\,1.0 & 27.4\,$\pm$\,0.2 & \textbf{45.2\,$\pm$\,0.4} \\
 & PPO $\alpha=1.5$ & 43.2\,$\pm$\,0.4 & 60.9\,$\pm$\,0.2 & 81.5\,$\pm$\,0.3 & 35.8\,$\pm$\,0.6 & 16.4\,$\pm$\,0.5 & 27.2\,$\pm$\,0.6 & 44.3\,$\pm$\,0.3 \\
\addlinespace
64 & GRPO & 45.9\,$\pm$\,0.2 & 61.8\,$\pm$\,0.3 & 82.5\,$\pm$\,0.3 & 35.5\,$\pm$\,1.1 & 16.7\,$\pm$\,0.4 & 28.1\,$\pm$\,0.1 & 44.9\,$\pm$\,0.3 \\
 & PPO $\alpha=0.1$ & 47.2\,$\pm$\,0.6 & 62.6\,$\pm$\,0.3 & \textbf{83.1\,$\pm$\,0.2} & 37.3\,$\pm$\,0.4 & \textbf{18.1\,$\pm$\,0.1} & 28.4\,$\pm$\,0.6 & \textbf{45.9\,$\pm$\,0.2} \\
 & PPO $\alpha=0.5$ & \textbf{47.9\,$\pm$\,0.1} & 63.2\,$\pm$\,0.2 & 82.1\,$\pm$\,0.5 & \textbf{37.4\,$\pm$\,0.8} & 17.0\,$\pm$\,0.1 & 28.7\,$\pm$\,0.4 & 45.7\,$\pm$\,0.2 \\
 & PPO $\alpha=1.5$ & 47.3\,$\pm$\,0.6 & \textbf{63.3\,$\pm$\,0.1} & 83.0\,$\pm$\,0.2 & 36.8\,$\pm$\,1.3 & 17.2\,$\pm$\,0.3 & \textbf{28.8\,$\pm$\,0.2} & 45.8\,$\pm$\,0.3 \\
\addlinespace
256 & GRPO & 50.2\,$\pm$\,0.4 & 63.5\,$\pm$\,0.0 & 83.7\,$\pm$\,0.5 & 38.0\,$\pm$\,0.4 & 17.6\,$\pm$\,0.2 & 29.5\,$\pm$\,0.3 & 46.5\,$\pm$\,0.1 \\
 & PPO $\alpha=0.1$ & 50.7\,$\pm$\,0.2 & 63.6\,$\pm$\,0.2 & 83.4\,$\pm$\,0.4 & 36.9\,$\pm$\,1.5 & 17.2\,$\pm$\,0.2 & 29.1\,$\pm$\,0.4 & 46.0\,$\pm$\,0.3 \\
 & PPO $\alpha=0.5$ & \textbf{52.3\,$\pm$\,0.5} & \textbf{64.7\,$\pm$\,0.6} & \textbf{84.3\,$\pm$\,0.4} & 38.6\,$\pm$\,0.6 & 17.7\,$\pm$\,0.3 & \textbf{31.0\,$\pm$\,0.4} & \textbf{47.3\,$\pm$\,0.1} \\
 & PPO $\alpha=1.5$ & 51.3\,$\pm$\,0.1 & 63.7\,$\pm$\,0.3 & 83.5\,$\pm$\,0.4 & \textbf{39.9\,$\pm$\,1.2} & \textbf{18.0\,$\pm$\,0.4} & 30.8\,$\pm$\,0.3 & 47.2\,$\pm$\,0.4 \\
\addlinespace
1{,}024 & GRPO & 55.1\,$\pm$\,0.5 & \textbf{66.2\,$\pm$\,0.3} & 84.4\,$\pm$\,0.1 & 41.0\,$\pm$\,0.7 & \textbf{18.1\,$\pm$\,0.2} & \textbf{31.8\,$\pm$\,0.1} & \textbf{48.3\,$\pm$\,0.1} \\
 & PPO $\alpha=0.1$ & 52.6\,$\pm$\,0.2 & 63.8\,$\pm$\,0.5 & 83.8\,$\pm$\,0.4 & 37.1\,$\pm$\,0.8 & 17.1\,$\pm$\,0.3 & 29.9\,$\pm$\,0.3 & 46.3\,$\pm$\,0.2 \\
 & PPO $\alpha=0.5$ & \textbf{55.6\,$\pm$\,0.2} & 65.0\,$\pm$\,0.3 & 84.5\,$\pm$\,0.3 & \textbf{41.4\,$\pm$\,0.4} & 17.9\,$\pm$\,0.3 & 31.6\,$\pm$\,0.4 & 48.1\,$\pm$\,0.1 \\
 & PPO $\alpha=1.5$ & 55.0\,$\pm$\,0.3 & 65.5\,$\pm$\,0.2 & \textbf{85.1\,$\pm$\,0.3} & 39.6\,$\pm$\,0.4 & 17.5\,$\pm$\,0.2 & 31.7\,$\pm$\,0.3 & 47.9\,$\pm$\,0.1 \\
\addlinespace
4{,}096 & GRPO & 56.8\,$\pm$\,0.2 & 65.8\,$\pm$\,0.4 & 84.8\,$\pm$\,0.5 & 40.2\,$\pm$\,0.4 & \textbf{17.6\,$\pm$\,0.1} & 32.2\,$\pm$\,0.4 & 48.1\,$\pm$\,0.2 \\
 & PPO $\alpha=0.1$ & 54.1\,$\pm$\,0.2 & 64.1\,$\pm$\,0.3 & 84.2\,$\pm$\,0.3 & 39.3\,$\pm$\,0.7 & 17.4\,$\pm$\,0.3 & 29.4\,$\pm$\,0.4 & 46.9\,$\pm$\,0.2 \\
 & PPO $\alpha=0.5$ & 56.9\,$\pm$\,0.3 & \textbf{66.5\,$\pm$\,0.1} & \textbf{85.1\,$\pm$\,0.2} & 40.5\,$\pm$\,0.7 & 17.5\,$\pm$\,0.3 & 31.5\,$\pm$\,0.2 & \textbf{48.2\,$\pm$\,0.2} \\
 & PPO $\alpha=1.5$ & \textbf{57.5\,$\pm$\,0.2} & 65.6\,$\pm$\,0.1 & 84.5\,$\pm$\,0.3 & \textbf{40.9\,$\pm$\,0.2} & 16.7\,$\pm$\,0.1 & \textbf{32.6\,$\pm$\,0.2} & 48.1\,$\pm$\,0.1 \\
\midrule
-- & Qwen2.5-3B (base) & 15.3 & 29.2 & 42.0 & 12.7 & 5.8 & 10.2 & 20.0 \\
\bottomrule
\end{tabular}
\caption{\textbf{Full math results.} Scores at each run's best checkpoint
within 400 steps (mean $\pm$ one standard error over three seeds, accuracy in
\%).
DeepScaleR selects checkpoints. Avg.\ is the five-benchmark test average
in Figure~\ref{fig:short-math-best}. Bold marks the highest mean at each
batch size. The last row is the base model.}
\label{tab:math-all-scores}
\end{table*}

\begin{table*}[t]
\centering

\footnotesize
\begin{tabular}{@{}llcccc@{}}
\toprule
$B$ & Method & TMax-15K & TB 2.1 & TB-Lite & Avg. \\
\midrule
128 & GRPO & 52.7\,$\pm$\,0.4 & 17.3\,$\pm$\,0.8 & 40.9\,$\pm$\,1.0 & 29.1\,$\pm$\,0.7 \\
 & PPO $\alpha=0.5$ & 56.1\,$\pm$\,0.4 & \textbf{19.0\,$\pm$\,1.2} & \textbf{48.6\,$\pm$\,1.4} & \textbf{33.8\,$\pm$\,0.9} \\
 & PPO $\alpha=1.5$ & \textbf{58.1\,$\pm$\,0.9} & 17.3\,$\pm$\,0.4 & 46.7\,$\pm$\,1.0 & 32.0\,$\pm$\,0.5 \\
\addlinespace
512 & GRPO & 61.5\,$\pm$\,0.1 & 19.2\,$\pm$\,0.6 & 46.6\,$\pm$\,0.9 & 32.9\,$\pm$\,0.5 \\
 & PPO $\alpha=0.5$ & \textbf{66.6\,$\pm$\,0.2} & \textbf{23.6\,$\pm$\,0.6} & 51.3\,$\pm$\,1.5 & 37.5\,$\pm$\,0.9 \\
 & PPO $\alpha=1.5$ & 62.3\,$\pm$\,0.5 & 22.3\,$\pm$\,1.1 & \textbf{55.2\,$\pm$\,1.0} & \textbf{38.8\,$\pm$\,0.6} \\
\addlinespace
2{,}048 & GRPO & \textbf{70.8\,$\pm$\,0.8} & 22.6\,$\pm$\,1.2 & 55.2\,$\pm$\,0.7 & 38.9\,$\pm$\,0.6 \\
 & PPO $\alpha=0.5$ & 69.7\,$\pm$\,0.8 & 21.6\,$\pm$\,0.9 & \textbf{56.5\,$\pm$\,1.4} & \textbf{39.1\,$\pm$\,0.9} \\
 & PPO $\alpha=1.5$ & 69.5\,$\pm$\,0.4 & \textbf{22.8\,$\pm$\,1.1} & 54.3\,$\pm$\,1.2 & 38.5\,$\pm$\,0.7 \\
\midrule
-- & Qwen3.5-4B (initial) & 35.0 & 9.6 & 26.4 & 18.0 \\
-- & TMax-4B & -- & 16.7 & 43.8 & 30.3 \\
\bottomrule
\end{tabular}
\caption{\textbf{Full terminal-agent results.} Scores with one training seed
(mean $\pm$ one standard error over evaluations, in \%). All columns use the final checkpoint (step 200).
TMax-15K is the validation set, scored with avg@3. Terminal-Bench 2.1 (TB 2.1) and TB-Lite
report mean reward over eight
attempts per task, and Avg.\ is their mean.
Bold marks the highest mean
at each batch size. TMax-4B is the released Tmax
model~\citep{ivison2026tmax}.}
\label{tab:terminal-all-scores}
\end{table*}

\subsection{Other Control Methods}
\label{app:all-control}
\label{app:std-normalization}

\begin{figure*}[t]
\centering
\includegraphics{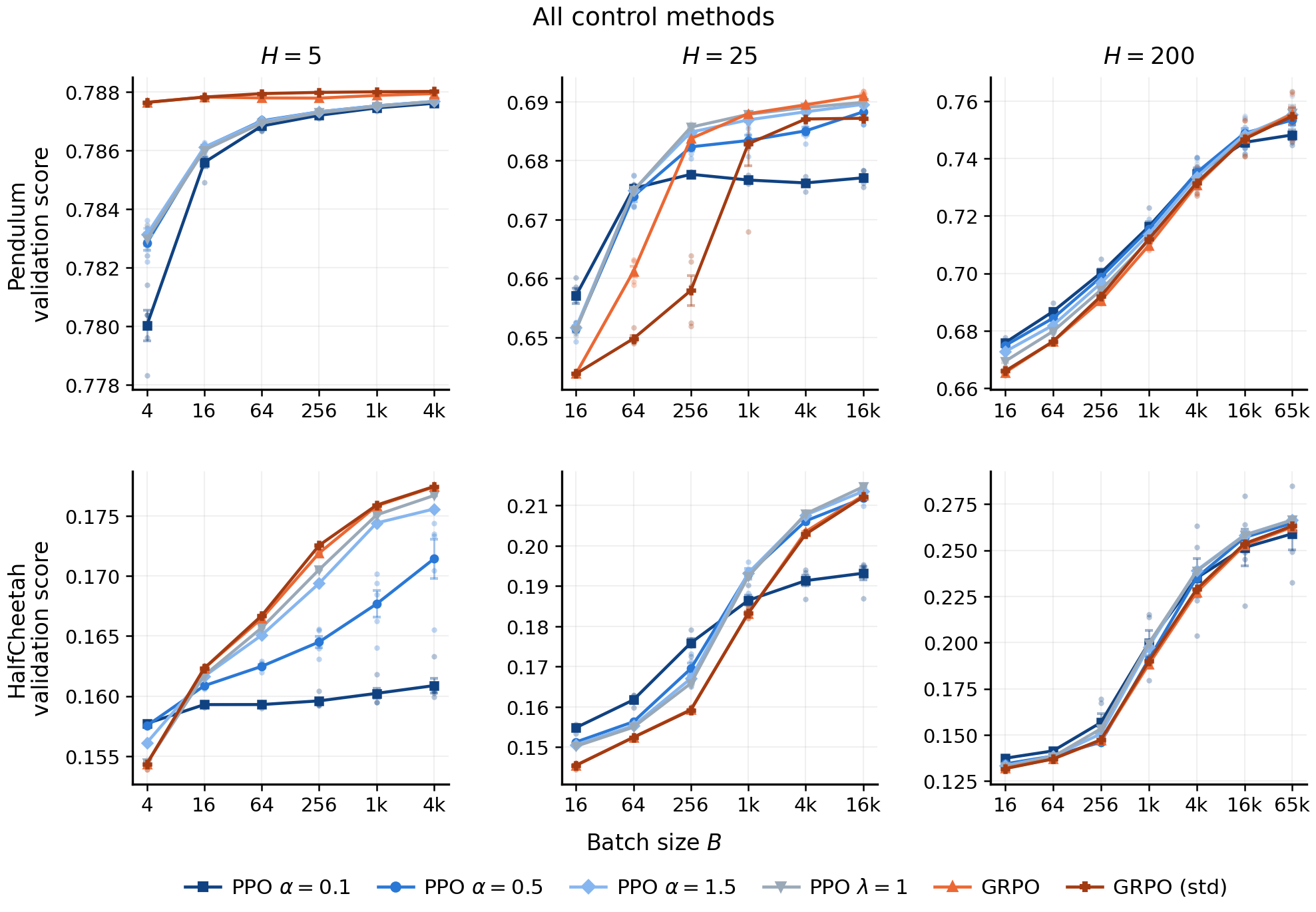}
\caption{\textbf{All control methods.} Best validation score for every control method, as in
Figure~\ref{fig:gymnasium-batch-scaling} with PPO $\lambda=1$ and GRPO (std)
added.}
\label{fig:gymnasium-all}
\end{figure*}

\begin{figure*}[t]
\centering
\includegraphics{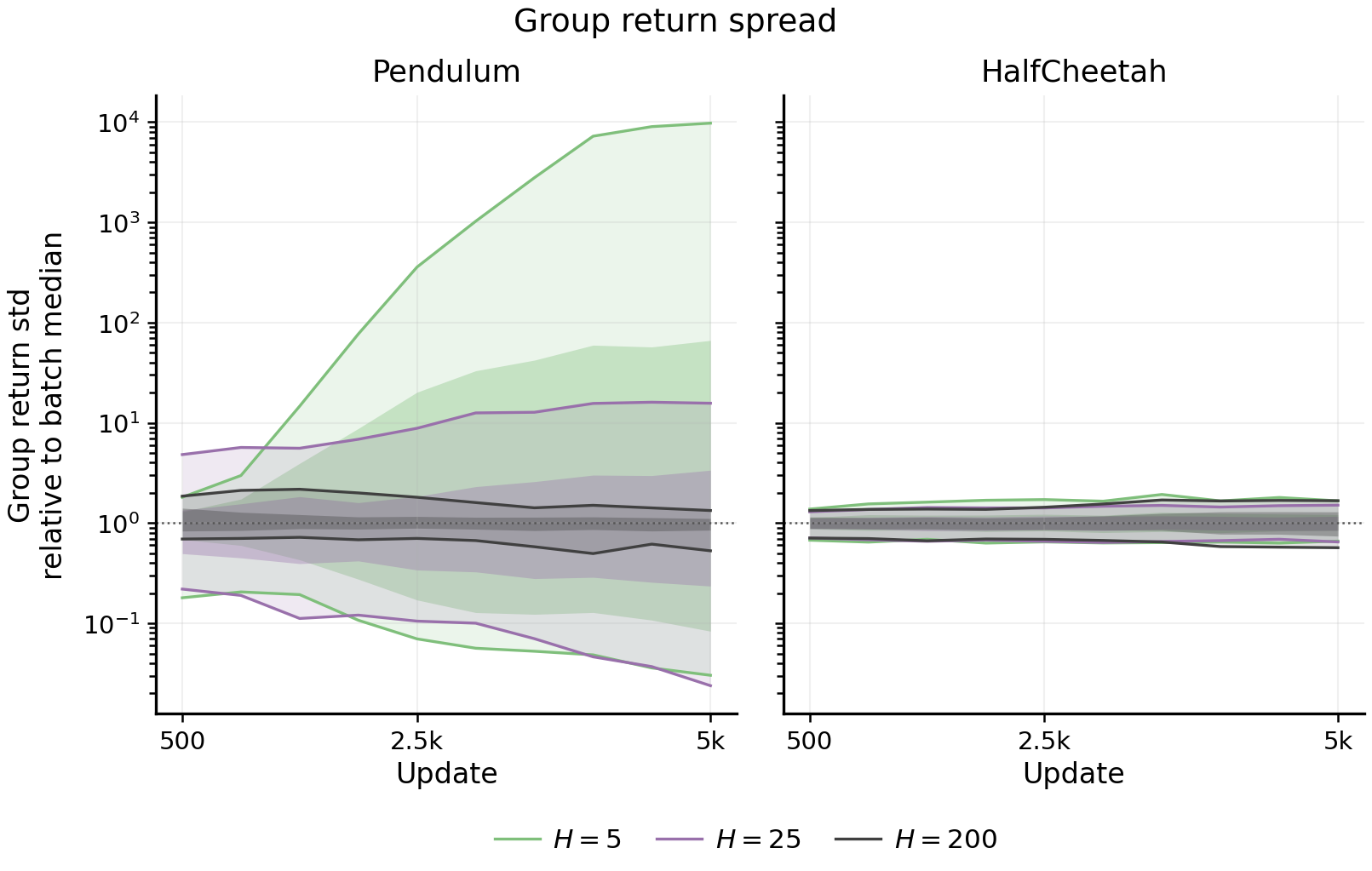}
\caption{\textbf{Group return spread in control tasks.} Group return standard deviation relative to the batch median over
training, for the GRPO run with $B=1{,}024$. Dark bands span the 25th to 75th
percentiles and light bands the 5th to 95th.}
\label{fig:group-std-spread}
\end{figure*}

Figure~\ref{fig:gymnasium-all} adds two methods to
Figure~\ref{fig:gymnasium-batch-scaling}. PPO with $\lambda=1$, which does not
bootstrap, has a mean score within $0.004$ of PPO with $\alpha=1.5$ at every
batch size. GRPO
(std) normalizes each group's advantages by its standard deviation. Its mean
score is within $0.003$ of GRPO at every batch size except in Pendulum at
$H=25$.

With binary rewards and $G\le16$, as in math, a group with mixed outcomes has a reward
standard deviation of at least $0.24$, so normalization rescales groups by a
bounded factor. Continuous returns have no such floor.
Figure~\ref{fig:group-std-spread} shows the spread of group standard
deviations along the GRPO run with $B=1{,}024$, measured with 16 trajectories
from each of 512 start states every 500 updates. In HalfCheetah and in
Pendulum at $H=200$, the 95th percentile stays within about $3\times$ the 5th,
so normalization rescales groups by similar factors. In Pendulum at $H=25$, the
ratio grows from $22$ to $657$ over training, and at $H=5$ it reaches
$3\times10^5$. Normalization gives every group advantages of unit scale, so
the start states whose returns barely vary receive as much weight as those with
the most to learn. In Pendulum at $H=5$, both variants score within $0.001$ of their
$B=4{,}096$ score already at $B=4$, so the reweighting costs nothing. At
$H=25$, scores still improve with $B$, and GRPO (std) falls behind.

\subsection{Privileged and Unprivileged Critics}
\label{app:priv-ablation}

\begin{figure*}[t]
\centering
\includegraphics{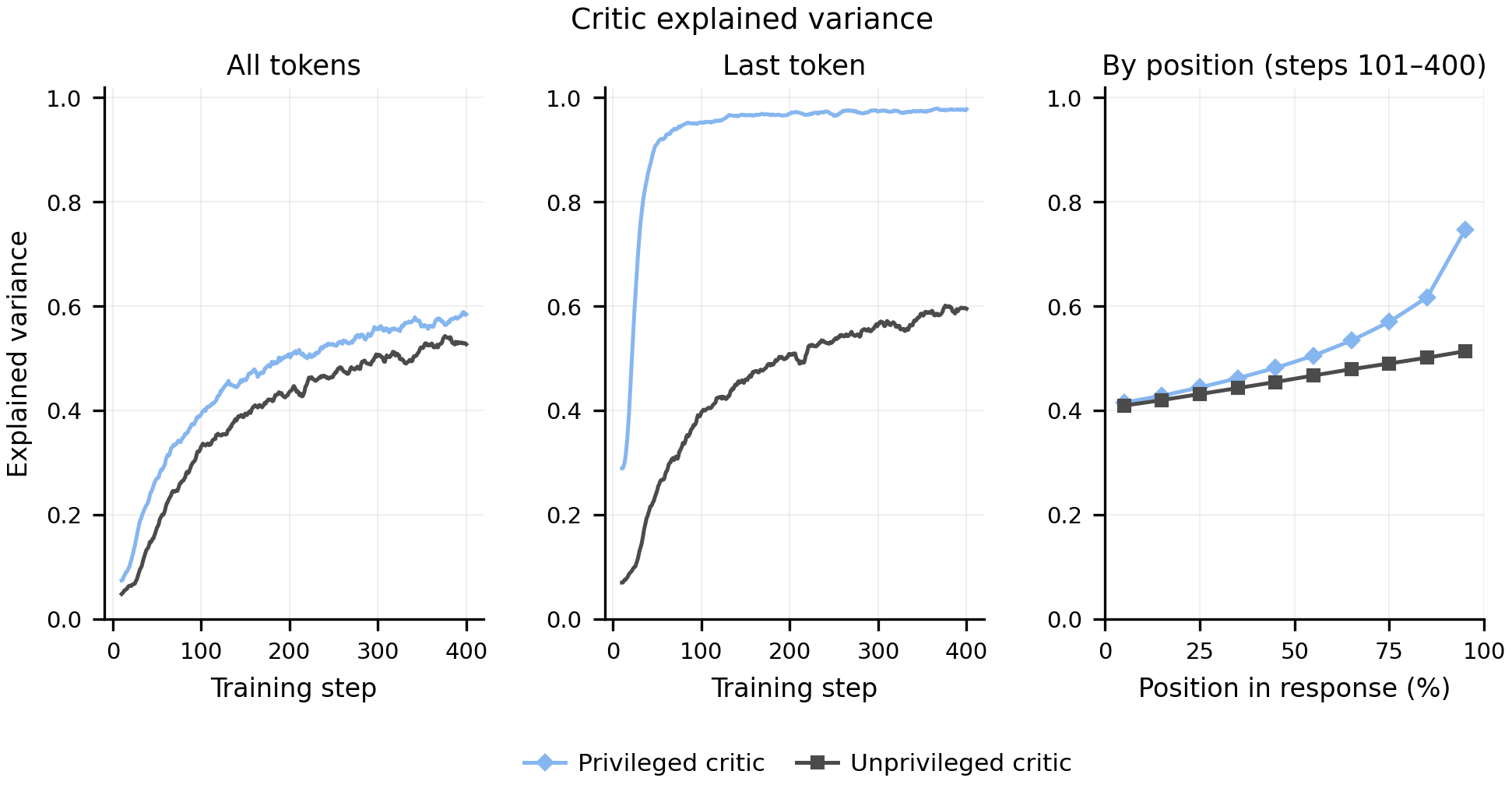}
\caption{\textbf{Critic explained variance in math.} Critic explained variance for PPO with $\alpha=1.5$ and $B=1{,}024$ in math (one seed), over all policy tokens (left) and at the last token (middle) as 10-step
rolling means, and by position in the response over steps 101--400 (right).}
\label{fig:priv-ev}
\end{figure*}

\begin{figure}[t]
\centering
\includegraphics{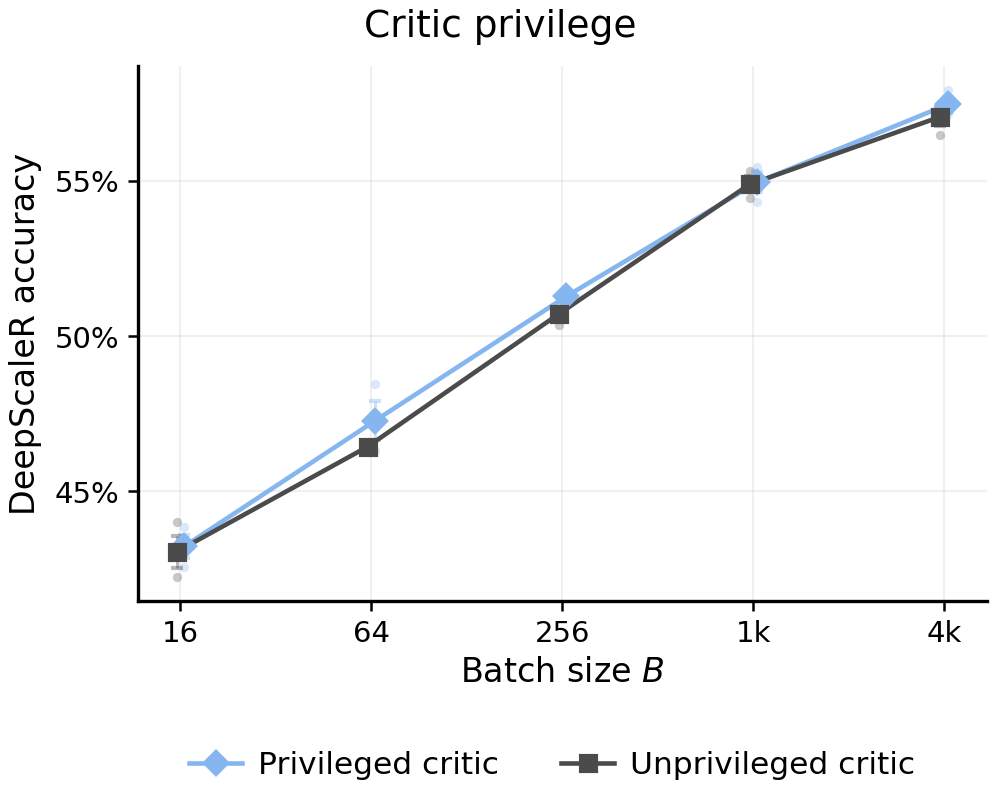}
\caption{\textbf{Privileged vs.\ unprivileged critic.} Best DeepScaleR
accuracy within 400 steps for PPO with $\alpha=1.5$ and a privileged or
unprivileged critic. Mean and one standard error over three seeds.}
\label{fig:priv-ablation}
\end{figure}

The privileged critic explains more of the return variance than an unprivileged critic that observes only the actor's
inputs, and the gap grows along the response (Figure~\ref{fig:priv-ev}). The
two critics are nearly identical near the start of a response, while at the
last token the privileged critic explains $0.97$ of the variance after step
100, against $0.53$. Accuracy is similar. On DeepScaleR validation, the
two critics differ by at most $0.9$ points at every tested batch size, from
$16$ to $4{,}096$ (Figure~\ref{fig:priv-ablation}), and they are within $0.3$ points on the
five-benchmark average. 

\clearpage
\section{Gradient Measurements}
\label{app:llm-gradient}

This appendix details the control-task gradient measurements of
Section~\ref{sec:llm-gradient-results} and the group-size check of
Section~\ref{sec:comparison-protocol}. Each measurement
compares an estimator's batch gradient $\hat g_B$ with a reference estimate of
the policy gradient $g$ at a frozen checkpoint.

\subsection{Control Tasks: Method}
\label{app:control-gradient}
\label{app:pendulum-training}

\paragraph{Checkpoints.}
In HalfCheetah, for each horizon $H\in\{5,25,200\}$ and each PPO variant, we
use that variant's training run with $B=1{,}024$ (seed 1), with its actor and
critic, every 500 updates. GRPO has no critic, so we evaluate its estimator
on the same policies as each PPO variant. In Pendulum, runs at $B=1{,}024$
reach their score plateau after a median of about 1,200 updates at $H=5$ and
2,300 at $H=25$, after which $\|g\|$ shrinks and $\nu$ rises sharply. We
therefore measure Pendulum on runs that are still improving, trained at
$B=16$, $64$, and $1{,}024$ at $H=5$, $25$, and $200$, respectively. A run's score plateau is
the first update at which a centered 201-update moving average of its
training score reaches 80\% of its total gain.

\paragraph{Reference gradient.}
We draw a panel of 256 start states from the training start distribution with
a fixed seed. For each state, we sample 8,192 trajectories in two halves of
4,096 and compute a leave-one-out policy gradient per state and half, using the
score of the pre-$\tanh$ Gaussian action. The reference is the mean over
states and halves, with reference noise removed using the two halves. We
compute exact per-trajectory gradients over all actor parameters (5,772 in
HalfCheetah), without sketching. Across all checkpoints, the halves agree with
cosine at least $0.989$ in Pendulum and $0.986$ in HalfCheetah, so the
reference has cosine at least $0.996$ with the panel's true gradient.

\paragraph{Estimators and batches.}
We draw 128 independent pools, each with 256 fresh trajectories from every
panel state, and take up to 16 disjoint batches from each pool at each batch
size. GRPO subtracts the group mean without standard-deviation normalization,
and PPO uses GAE with the checkpoint critic and length-adaptive $\lambda$
(Equation~\ref{eq:length-adaptive-gae}, floored at $0$), whitened over the $BH$
transitions of the batch, as in training. Both estimators use the same batches
of $P=B/16$ groups of $G=16$ trajectories, for $B$ from 16 to 65,536, so they
differ only in their advantages. Up to $B=4{,}096$, each group has a different
state, and above it each of the 256 states supplies $B/4{,}096$ groups, each
with its own trajectories. PPO trains with $G=1$, which would need more
distinct states than the panel holds. In HalfCheetah, PPO's expected cosine at
$G=16$ is within $0.01$ of its value at $G=1$ at every batch size of the
group-size check (Appendix~\ref{app:group-size}).

\paragraph{Metrics.}
We report the mean cosine to the reference, whose noise is negligible here.
Both $c$ and $\nu$ come from the 128 batches of $B=65{,}536$, one per pool. The
alignment $c$ is the cosine of their mean with the reference, and $\nu$ is $B$
times their normalized variance. The standard error of $c$ comes from leaving
out each of eight groups of pools in turn.

\subsection{Cosine Model, Horizon Bound, and Noise Scale}
\label{app:control-checks}

\begin{figure*}[t]
\centering
\includegraphics{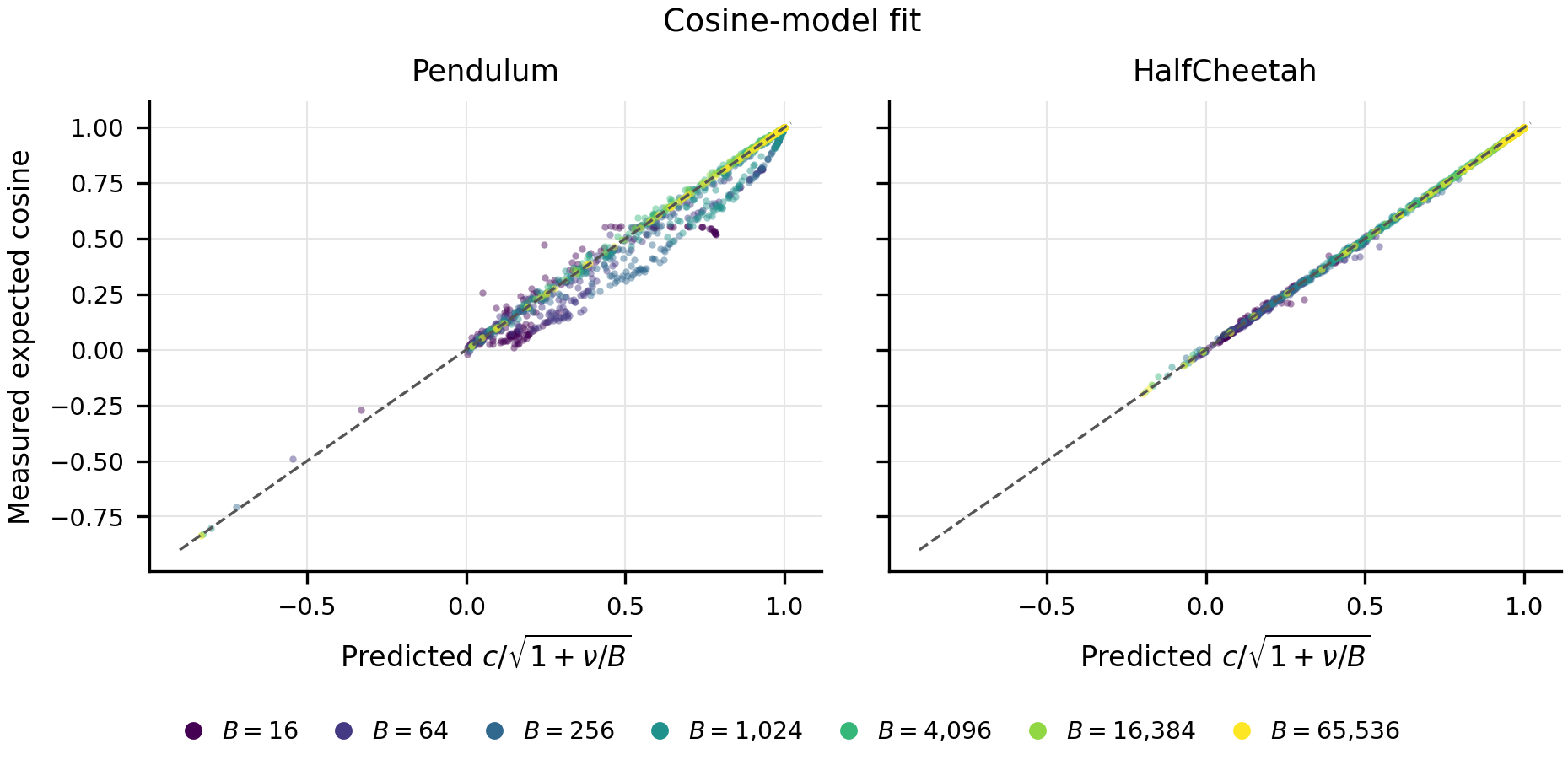}
\caption{\textbf{Cosine-model fit across checkpoints.} Each
point is one checkpoint, estimator, and batch size from updates 0 to 5,000,
with 1,302 points per environment. At $B<65{,}536$, where the prediction is out
of sample, the median absolute error is $0.015$ in Pendulum and $0.003$ in
HalfCheetah. The dashed line is equality.}
\label{fig:eq7-fit}
\end{figure*}

\begin{table*}[t]
\centering
\small
\setlength{\tabcolsep}{6pt}
\renewcommand{\arraystretch}{1.15}
\begin{tabular*}{0.8\textwidth}{@{\extracolsep{\fill}}lrrrrrr@{}}
\toprule
Environment & $H$ & Training $B$ & GRPO $\nu$ & $C$ & $\|g\|^2$ & $V/(CH)$ \\
\midrule
Pendulum & 5 & 16 & 13 & 13 & $2.0\times10^{-4}$ & 0.55 \\
 & 25 & 64 & 447 & 21 & $1.1\times10^{-3}$ & 0.36 \\
 & 200 & 1{,}024 & 533 & 42 & $6.0\times10^{-3}$ & 0.36 \\
\addlinespace
HalfCheetah & 5 & 1{,}024 & 144 & 100 & $2.0\times10^{-4}$ & 0.018 \\
 & 25 & 1{,}024 & 1{,}949 & 120 & $3.0\times10^{-4}$ & 0.020 \\
 & 200 & 1{,}024 & 4{,}032 & 105 & $4.0\times10^{-4}$ & 0.018 \\
\bottomrule
\end{tabular*}
\caption{\textbf{Gradient statistics by environment and horizon.} Geometric
means over checkpoints from update 500 until each run's score plateau. $C$ is
the largest per-step expected squared score norm, as in
Proposition~\ref{cor:outcome-gradient-variance}, and $V$ is the variance of the
Monte Carlo estimator of Equation~\ref{eq:mc-trajectory-gradient}.}
\label{tab:control-stats}
\end{table*}

\paragraph{Cosine model.}
Figure~\ref{fig:eq7-fit} covers GRPO and each run's own PPO estimator. The
median absolute error falls from $0.047$ at $B=16$ to $0.006$ at
$1{,}024\le B\le16{,}384$ in Pendulum, and from $0.004$ to $0.002$ in
HalfCheetah, since the squared norm of the batch gradient concentrates less at
small batch sizes (Equation~\ref{eq:cosine-error}).

\paragraph{Horizon bound.}
At each checkpoint, we draw 4,096 fresh trajectories, 16 from each panel state,
and compute the Monte Carlo estimator of
Equation~\ref{eq:mc-trajectory-gradient} with the uncentered score $S(\tau)$ as
reward. We estimate $C$ as
the largest per-step $\mathbb E\|\nabla_\theta\log\pi_\theta(a_t\mid s_t)\|^2$
and $V$ as $\mathbb E\|\hat g\|^2-\|g\|^2$. Both bounds of
Equation~\ref{eq:horizon-batch-variance} hold at every checkpoint. The bound's slack, shown by $V/(CH)$ in Table~\ref{tab:control-stats}, comes mostly from rewards
below one.

\paragraph{Noise-scale measurement.}
Without a baseline, every trajectory's gradient is scaled by its full reward,
so the Monte Carlo estimator's noise scale mostly reflects the mean reward. At
Pendulum's initial policy with $H=5$, it is about $7\times10^{4}$, against $10$
for GRPO. GRPO's noise scale also does not change when rewards are shifted or rescaled,
whereas PPO's reflects its critic's error, which differs between environments.
We summarize each run by the geometric mean over checkpoints from update $500$
until its score plateaus. Later checkpoints are excluded because the noise scale
is inflated near convergence, where the true gradient becomes small. Pendulum
and HalfCheetah also differ in dynamics and rewards, so the domain comparison in
Section~\ref{sec:llm-gradient-results} does not isolate dimensionality.

\subsection{Group Size}
\label{app:group-size}

\begin{figure*}[t]
\centering
\includegraphics[width=\textwidth]{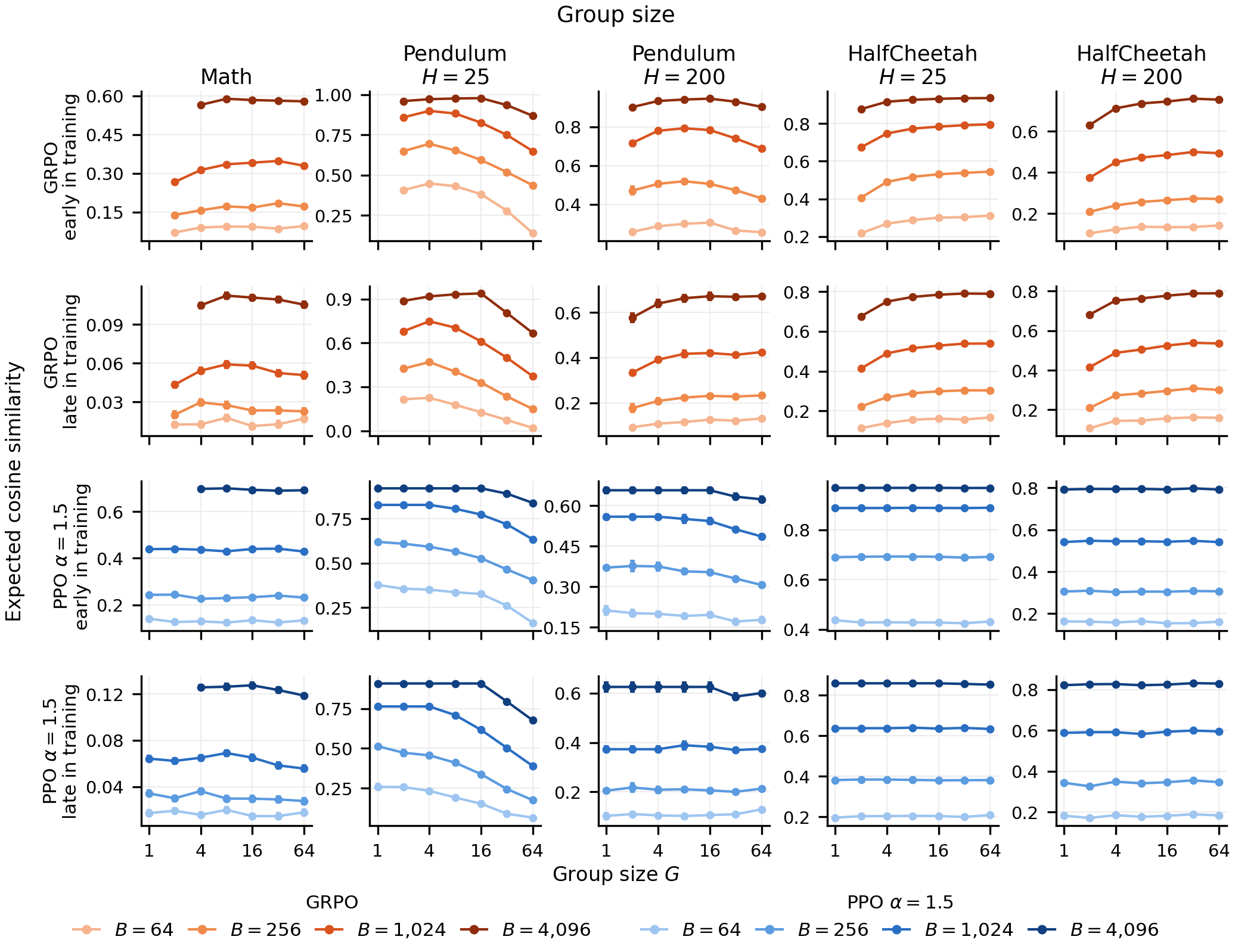}
\caption{\textbf{Expected cosine against group size.} Cosine to the policy
gradient against group size $G$ for GRPO and PPO with $\alpha=1.5$, with one
line per batch size $B$. Early in training is step 0 in math and update 500 in
control tasks, and late in training is step 140 and update 3,000. Math uses GRPO with
standard-deviation normalization and control tasks use Dr.~GRPO, as in training.
Error bars show one standard error over batches.}
\label{fig:group-size}
\end{figure*}

Figure~\ref{fig:group-size} varies the group size at fixed batch size for GRPO
and PPO with $\alpha=1.5$. In math, we measure steps 0 and 140 of the PPO $\alpha=1.5$ run with
$B=1{,}024$, following an existing sketched-gradient
methodology~\citep{anonymous2026measuring}. Per-response gradients over all model
parameters are compressed with CountSketch~\citep{charikar2002countsketch} (65,536 buckets)
and compared with a
reference estimated from a larger, disjoint set of responses on a panel of
1,024 DeepScaleR training questions, using independent halves to remove
sampling noise from norms and cosines. In control tasks, we use
the exact measurement of Appendix~\ref{app:control-gradient} at updates 500
and 3,000 of each $\alpha=1.5$ run, with Pendulum at $H=25$ and $H=200$,
trained at $B=64$ and $B=1{,}024$, and HalfCheetah at both horizons, trained at
$B=1{,}024$. Each batch splits into $B/G$ prompts or start states with $G$
trajectories each, reusing control start states when $B/G>256$. Math batches cannot exceed the 1,024 distinct prompts of
its panel, so $G<4$ is not measured at $B=4{,}096$.

PPO's expected cosine changes little with group size or is highest at small
$G$ in every setting, and at $B=4{,}096$ its smallest measured group size is within $0.01$
of its best. For GRPO, $G=16$ is within $0.015$ of the best group size at
$B=4{,}096$, the largest batch size in this check, in every setting, and within
$0.02$ at every batch size in math and HalfCheetah. Where $G=16$ is not GRPO's
best, a better group size could only raise GRPO's expected cosine, so the
batch sizes at which GRPO crosses over are, if anything, conservative.

\end{document}